\documentclass[10pt,twocolumn,letterpaper]{article}

\usepackage[pagenumbers]{wacv} 

\usepackage{indentfirst}
\usepackage{multirow}
\usepackage{bm}
\usepackage{makecell}
\usepackage{microtype}

\definecolor{wacvblue}{rgb}{0.21,0.49,0.74}
\usepackage[pagebackref,breaklinks,colorlinks,allcolors=wacvblue]{hyperref}

\def\wacvPaperID{340} 
\def\confName{WACV}
\def\confYear{2026}

\title{ViGG: Robust RGB-D Point Cloud Registration using Visual-Geometric Mutual Guidance}

\author{Congjia Chen\\
Beihang University\\
{\tt\small sy2317314@buaa.edu.cn}
\and
Shen Yan\\
China Agricultural University\\
{\tt\small 2022310030201@cau.edu.cn}
\and
Yufu Qu\\
Beihang University\\
{\tt\small qyf@buaa.edu.cn}
}

\begin{document}
\maketitle
\begin{abstract}
\indent Point cloud registration is a fundamental task in 3D vision. Most existing methods only use geometric information for registration. Recently proposed RGB-D registration methods primarily focus on feature fusion or improving feature learning, which limits their ability to exploit image information and hinders their practical applicability. In this paper, we propose ViGG, a robust RGB-D registration method using mutual guidance. First, we solve clique alignment in a visual-geometric combination form, employing a geometric guidance design to suppress ambiguous cliques. Second, to mitigate accuracy degradation caused by noise in visual matches, we propose a visual-guided geometric matching method that utilizes visual priors to determine the search space, enabling the extraction of high-quality, noise-insensitive correspondences. This mutual guidance strategy brings our method superior robustness, making it applicable for various RGB-D registration tasks. The experiments on 3DMatch, ScanNet and KITTI datasets show that our method outperforms recent state-of-the-art methods in both learning-free and learning-based settings. Code is available at \href{https://github.com/ccjccjccj/ViGG}{\textcolor[rgb]{0.21,0.49,0.74}{https://github.com/ccjccjccj/ViGG}}.
\end{abstract}    
\section{Introduction}
\label{sec:intro}
Point cloud registration is important in 3D vision and serves as a foundation for downstream applications such as 3D reconstruction. The goal of registration is to estimate a rigid transformation between two point clouds, allowing their overlapping regions to be properly aligned.
\par
Feature-based methods are widely used in point cloud registration, they typically use hand-crafted descriptors\cite{fpfh, shot} or neural networks\cite{fcgf, predator} to extract local geometric features and find correspondences. Recent feature-based methods use a point convolutional backbone\cite{kpconv} to extract hierarchical features, and adopt attention modules to aggregate intra- and inter-frame information\cite{geotransformer, regtr, peal, dynamiccue}. However, ambiguous shapes and structures introduce global ambiguity to the geometric features. Even the most advanced networks have limited performance, and usually struggle to extract reliable correspondences in low-overlap scenarios. Recently, the popularization of RGB-D sensors facilitate registration methods using visual information\cite{byoc, llt, pointmbf, rgbdglue, colorpcr}. For instance, PointMBF\cite{pointmbf} extracts both visual and geometric features and fuses them through pixel-point mapping and bidirectional fusion. Similarly, ColorPCR\cite{colorpcr} generates color point cloud from RGB-D data and embeds point-wise color values into convolution layers to extract fused features. By fusing visual information, more distinctive features can be extracted. However, they utilize visual information in restricted ways and are not always applicable to various tasks. Differences in data acquisition lead to different spatial distributions of data, meaning there isn't always a valid pixel-point mapping for each image pixel or 3D point. This issue is particularly common in camera-LiDAR data, resulting in information loss. Additionally, when the mapping between pixels and 3D points is inaccurate, the fused features can be greatly impacted. These challenges limit their generalization, making it difficult for them to handle various registration tasks.

\begin{figure}[tp]
    \centering
    \includegraphics[width=\linewidth]{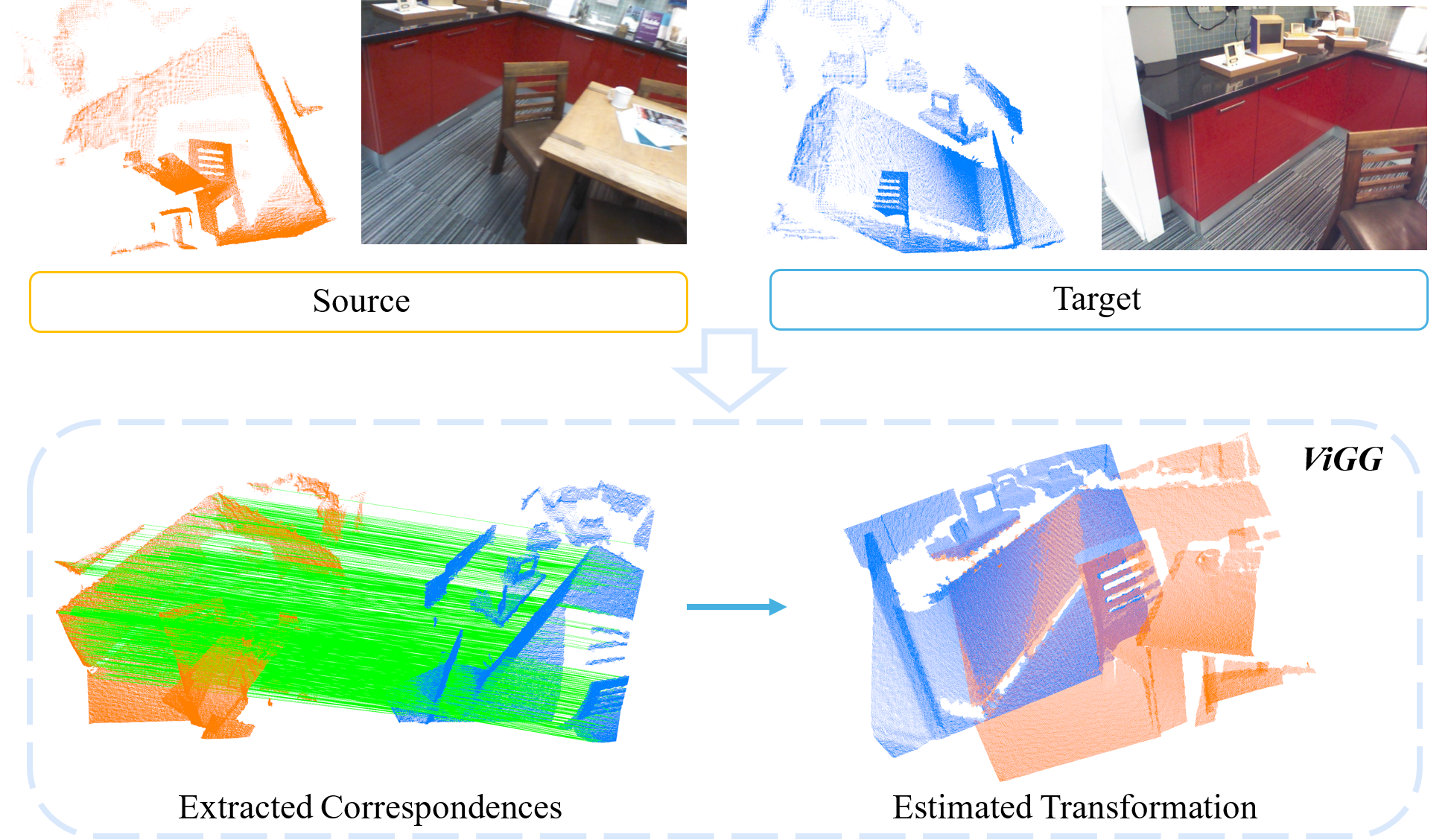}
    \caption{ViGG takes point clouds and their corresponding images as input, estimated the transformation between two point clouds.}
    \label{fig:sketch}
    \vspace{-1em}
\end{figure}

\par
For general applications, leveraging visual features extracted directly from images is promising. Compared to point cloud, the abundant texture information in image makes it easier to detect keypoints and describe features, leading to more reliable matching. Besides, existing image matching networks have shown remarkable performance and generalization after pretraining\cite{lightglue, omniglue}. However, registration based on visual matches can be hindered by ambiguous texture, and is highly sensitive to matching errors and 2D-to-3D mapping errors. Consequently, high-quality images and accurate calibration are necessary, which limits its application.
\par
In this paper, we propose ViGG, a robust RGB-D registration method using \textbf{Vi}sual-\textbf{G}eometric mutual \textbf{G}uidance. Unlike previous RGB-D registration methods that fuse multi-modality features using networks, we explore explicitly leveraging the complementary characteristics of geometric and visual matches, focusing on solving their inherent limitations in registration tasks. We first design a geometric-guided visual clique alignment module. This module takes advantage of the sparsity and high inlier ratio of visual matches, significantly boosting the efficiency of clique alignment. A geometric guidance design is also introduced to mitigate the ambiguity in visual cliques, which effectively improves robustness. To address the accuracy deterioration caused by noise in visual matches, we propose a visual-guided geometric matching module. By determining local search zones for geometric matching using visual guidance, the challenge of global ambiguity is sufficiently resolved, enabling extraction of high-quality correspondences. The incorporation of geometric features also makes the extracted correspondences insensitive to noise in visual matches, leading to accurate transformation estimation, as shown in \cref{fig:sketch}. 
\par
Compare to previous works, our mutual guidance strategy is simple and can sufficiently leverage the respective strengths of visual and geometric matches to overcome their limitations. Extensive experiments on indoor and outdoor datasets\cite{3dmatch, scannet, kitti} demonstrate our method's effectiveness on both learning-free and learning-based settings. Our method exhibits superior registration performance compared to recent methods, and demonstrates strong robustness in visually noisy cases. Our main contributions are as follows:
\par
$\bullet$ A robust point cloud registration method that takes images and point clouds as input, which is applicable to RGB-D sensor data and camera-LiDAR data.
\par
$\bullet$ A geometric-guided visual clique alignment method that solves maximal clique alignment in a new combination form, which effectively suppresses the ambiguous cliques in visual matches and has low computational cost.
\par
$\bullet$ A visual-guided geometric matching method that extracts high-quality correspondences with prior visual information, which effectively reduces the impact of noise in visual matches and improves the registration accuracy.
\section{Related Work}
\label{sec:related}

\noindent \textbf{Feature-based Point Cloud Registration.} Classical registration methods are typically ICP-based\cite{icp, gicp}, where the nearest neighbors are considered as matches, and the transformation is solved iteratively. In recent years, most methods focus on feature-based registration. They use hand-crafted descriptors\cite{fpfh, shot, 3dsc} or neural networks\cite{fcgf, ppfnet, predator, spinnet, parenet} to extract geometric features for each point, and match them based on feature similarity. CoFiNet\cite{cofinet} employs attention modules and coarse-to-fine strategy to improve feature matching. GeoTransformer\cite{geotransformer} introduces a geometric embedding to encode geometric structure for more distinctive features. PEAL\cite{peal} uses overlap prior to enhance feature extraction. Besides, several methods utilize color information to extract fused features\cite{llt, pointmbf, colorpcr}. In contrast, our method focus on leveraging extracted visual and geometric features for robust registration, rather than feature learning.
\par
\noindent \textbf{Robust Estimator.} Although feature-based methods can extract geometric features for matching, the low inlier ratio presents a significant challenge for transformation estimation. To address this issue, numerous approaches have been proposed to handle low inlier ratio\cite{fgr, teaser, sc2pcr, mac, tear}. RANSAC and its variants\cite{ransac, gcransac} are commonly used robust estimator methods, which can estimate correct transformation under heavy outliers. TEASER\cite{teaser} reformulates the registration problem using a Truncated Least Squares cost, making it insensitive to outliers. SC$^2$-PCR\cite{sc2pcr} introduces a second order spatial compatibility to identify inliers, while its improved version SC$^2$-PCR++\cite{sc2pcr++} proposes more reliable hypothesis evaluation metrics. MAC\cite{mac} constructs a compatibility graph and searches for maximal cliques to find the best transformation. Additionally, several works focus on training networks to find credible inliers\cite{3dregnet, dgr, pointdsc, vbreg}. Fuzzy clustering is also used for robust registration\cite{zhao2023accurate}, as it can handle inconsistent point-wise features which are common in cross-source registration. In comparison, we further explore robust registration using both geometric and visual information, which bring our method better performance.
\par
\noindent \textbf{Image Feature Matching.} To obtain image matches, traditional approaches detect keypoints and use hand-crafted descriptors\cite{sift, surf, orb} to extract local features, followed by matching through a nearest neighbor search. These traditional methods are still widely used in image matching tasks today. However, with the advancement of deep learning, learning-based methods have rapidly developed. Research focusing on Convolutional Neural Networks (CNNs)\cite{r2d2, d2net, superpoint} has demonstrated significant improvement in keypoint detection and feature description. Additionally, methods using attention modules have shown superior performance in keypoint matching\cite{superglue, sgmnet, lightglue, omniglue} and dense matching\cite{loftr, aspanformer}. The most recent sparse matching networks have exhibited remarkable generalization\cite{lightglue, omniglue}. Consequently, leveraging image matching methods for solving point cloud registration problem is promising, while most RGB-D registration methods focus on learning distinctive features through multi-modal learning\cite{byoc, llt, pointmbf, colorpcr}. In this paper, we explore directly leveraging visual matches for registration tasks.
\begin{figure*}[tp]
    \centering
    \includegraphics[width=\linewidth]{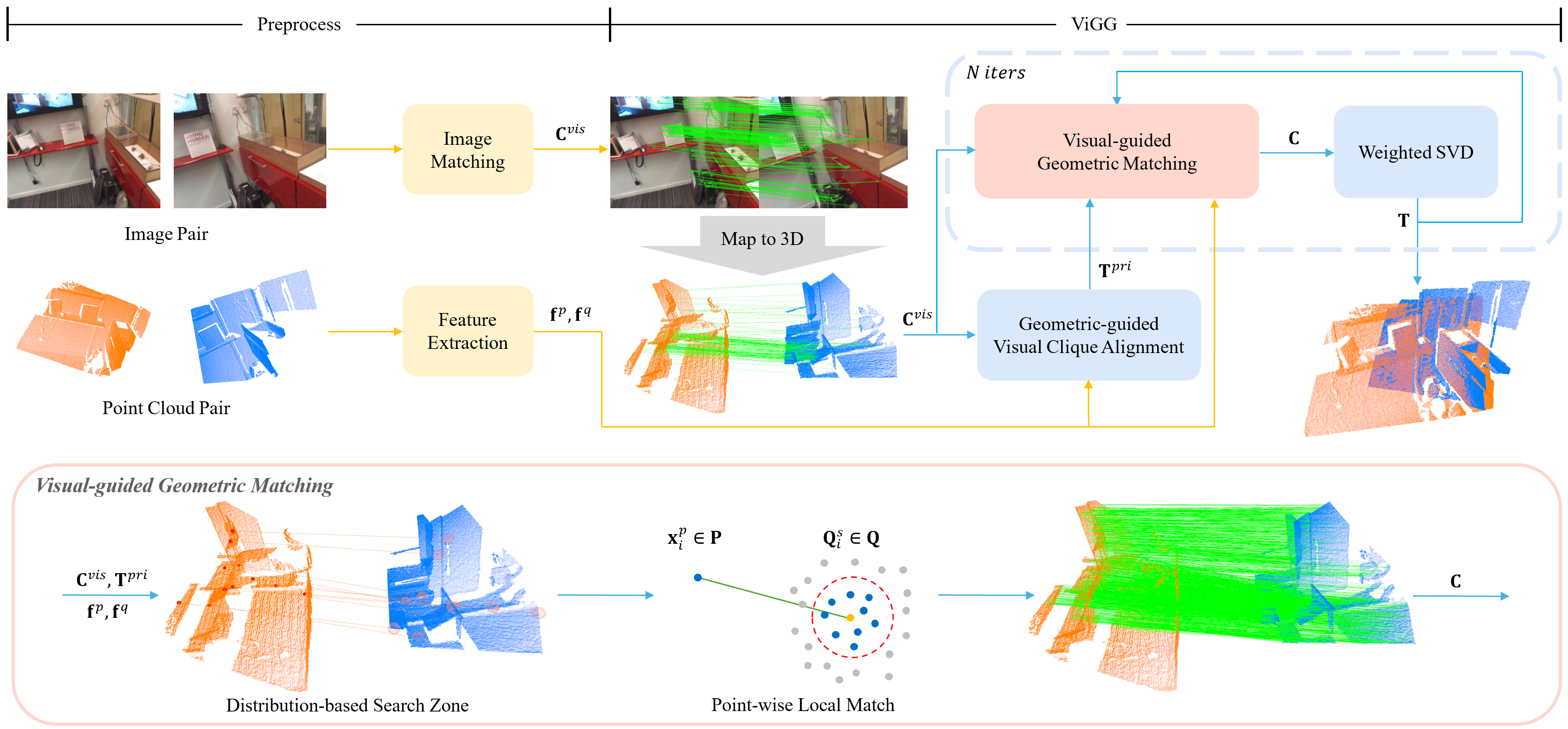}
    \caption{Pipeline of our method. Using the extracted visual matches and geometric features, the geometric-guided visual clique alignment module first estimates a prior transformation $\mathbf{T}^{pri}$. Then, the visual-guided geometric matching module iteratively determines the search zone for each point in $\mathbf{P}$ and extracts high-quality correspondences. The correspondences are used to estimate the updated transformation $\mathbf{T}$.}
    \label{fig:pipeline}
    \vspace{-1em}
\end{figure*}

\section{Method}
\label{sec:method}

\subsection{Problem Formulation}
To align two point clouds $\mathbf{P}$ and $\mathbf{Q}$, we first use visual descriptor to extract keypoint matches $\mathbf{C}^{vis}$ from their corresponding images. Using the given calibration information, we can determine the mapping between pixels and 3D points, and map the $\mathbf{C}^{vis}$ into 3D space. Then, we use geometric descriptor to extract local features for each point in two point clouds, and get two feature sets $\mathbf{f}^{p}, \mathbf{f}^{q}$. Unlike the pipelines of previous geometry-only or RGB-D registration methods, ViGG uses both visual match set $\mathbf{C}^{vis}$ and geometric features $\mathbf{f}^{p}, \mathbf{f}^{q}$ to estimate 6-DOF transformation between source $\mathbf{P}$ and target $\mathbf{Q}$, our pipeline is shown in \cref{fig:pipeline}.

\subsection{Geometric-guided Visual Clique Alignment}
Previous studies have demonstrated the effectiveness of clique alignment\cite{teaser, mac}. Inspired by this, we adapt the framework of MAC\cite{mac}. But different from MAC, we introduce a new combination form with a geometric guidance design to leverage the complementary characteristics of visual and geometric features. First, we provide a brief introduction of the core concept of MAC.
\par
MAC models the correspondences as a compatibility graph. Since inliers are geometrically compatible with each other, the graph can effectively determine them. To construct the compatibility graph, MAC first calculates the rigid distance difference between correspondences $\mathbf{c}_{i},\mathbf{c}_{j}$ in set $\mathbf{C}$:
\begin{equation} \label{eq1}
d_{ij} = \vert \Vert \mathbf{x}^{p}_{\mathbf{c}_{i}} - \mathbf{x}^{p}_{\mathbf{c}_{j}} \Vert_{2} - \Vert \mathbf{x}^{q}_{\mathbf{c}_{i}} - \mathbf{x}^{q}_{\mathbf{c}_{j}} \Vert_{2} \vert.
\end{equation}
Each correspondence is considered as a node in the graph, $\mathbf{c}_{i},\mathbf{c}_{j}$ will be linked with an edge if $d_{ij}$ is smaller than a threshold. After that, MAC searches the maximal cliques in the graph and finds the best transformation $\mathbf{T}^{pri}$ through hypothesis evaluation:
\begin{equation} \label{eq2}
\mathbf{T}^{pri} = \mathop{argmax}\limits_{\mathbf{T} \in h(\mathbf{C})} \sum^{\vert \mathbf{C} \vert}_{i=1}{g(\mathbf{c}_{i}, \mathbf{T})},
\end{equation}
$h(\mathbf{C})$ is the hypothesis set obtained from the maximal cliques in $\mathbf{C}$, $g$ is the error metric function formulated as:
\begin{equation} \label{eq3}
g(\mathbf{c}_{i}, \mathbf{T}) = max(0, \ t_{inlier} - \Vert \mathbf{T}(\mathbf{x}^{p}_{\mathbf{c}_{i}}) - \mathbf{x}^{q}_{\mathbf{c}_{i}} \Vert_{2}),
\end{equation}
where $\mathbf{T}(\mathbf{x})$ denotes applying transformation $\mathbf{T}$ to point $\mathbf{x}$ and $t_{inlier}$ is the inlier threshold.
\par
MAC is a strong estimator, however, finding maximal cliques takes exponential time in general and only can be quickly solved when graph is sparse\cite{teaser, arcs}. When the match set contains numerous inliers, the time and memory cost can be enormous. For this reason, the sparse but high-quality visual matches are inherently suitable for clique alignment. For geometric matching, finding repeatable keypoints is challenging, and even learning-based point cloud detectors struggle to achieve a high inlier ratio\cite{d3feat}, making extracting sparse correspondences difficult. But for visual matching, the texture information makes it easier to detect repeatable keypoints and extract distinctive features, which effectively control the size of graph while providing enough correct cliques. Benefited from the significantly fewer maximal cliques in graph, visual clique alignment can achieve a remarkable efficiency performance.
\par
While the efficiency is improved, solving clique alignment with visual matches meets two new challenges. First, visual matching can be easily affected by ambiguous textures or regions, and the ambiguous cliques supported by clustered incorrect visual matches might be mistakenly selected as the best hypothesis. Second, the accuracy of the estimated transformation is highly sensitive to noise in visual matches including matching errors and 2D-to-3D mapping errors, which significantly degrades the robustness. The second challenge will be discussed in the next section. To overcome the first challenge, we design a geometric guidance.
\par
By analyzing failure cases, we observe that some cases have established correct cliques, however, ambiguous cliques are supported by more incorrect matches and rank higher in hypothesis evaluation, which usually occurs in texture-similar regions. To address this, we perform global matching with geometric feature $\mathbf{f}^{p}, \mathbf{f}^{q}$ to obtain a geometric match set $\mathbf{C}^{geo}$, and introduce it into evaluation rather than using only visual matches, which reformulates the \cref{eq2} into:
\begin{equation} \label{eq4}
\mathbf{T}^{pri} = \mathop{argmax}\limits_{\mathbf{T} \in h(\mathbf{C}^{vis})} \sum^{\vert \mathbf{C'} \vert}_{i=1}{g(\mathbf{c}_{i}', \mathbf{T})},
\end{equation}
where $\mathbf{C'}=\mathbf{C}^{vis} \cup \mathbf{C}^{geo}$. The insight of this design is intuitive: inliers in geometric matches correspond to the correct transformation and will support the correct cliques, while outliers are irregular. When geometric matches contain sufficient inliers, the correct cliques can be clearly distinguished from ambiguous cliques. Conversely, when the inlier ratio is extremely low, since the search for maximal cliques is confined to visual matches, and geometric outliers are unlikely to support ambiguous visual cliques, which only occurs when visual ambiguous regions coincidentally coincide with geometric ambiguous regions. Therefore, geometric matches provide a loose guidance for hypothesis evaluation. They assist in identifying the correct cliques without introducing negative impacts when unreliable. Furthermore, this design is lightweight, only incurs negligible time cost. In the experiments, we will demonstrate that this simple design effectively overcomes the ambiguity in visual matches.

\subsection{Visual-guided Geometric Matching}
In common, there are localization and matching errors in visual matches. And due to rough calibration or different data distribution of sensors, the mapping between pixels and 3D points is not always accurate, which is particularly serious in LiDAR-camera data. Consequently, although the estimated $\mathbf{T}^{pri}$ is approximately correct, its accuracy is not promised. To address this limitation, we leverage geometric features to extract high-quality correspondences for accurate estimation. Geometric features would not be affected by visual noise or mapping error, however, local geometric features suffer from global ambiguity, which significantly limits the inlier ratio and presents a challenge to estimation. In this case, the inaccurate yet approximately correct $\mathbf{T}^{pri}$ can serve as a useful prior to guide geometric matching.
\par
\noindent \textbf{Error Distribution Estimation.} The first step is to evaluate the confidence of $\mathbf{T}^{pri}$. Unlike outliers, inliers can be precisely aligned under the ground truth transformation. If the transformation is inaccurate, it will introduce coordinate difference in the transformed inliers. Besides, due to matching errors, point resolution limitations or other factors, inliers in practice contain noise, which also results in coordinate difference. These two factors are interrelated, as noisy inliers often lead to less accurate transformation estimation. Therefore, the coordinate difference can be used to reflect the confidence of $\mathbf{T}^{pri}$. A large difference suggests that the transformation is likely to be inaccurate or the inliers are noisy, both of which imply that the estimation is not sufficiently reliable. We assume that the coordinate difference $\bm{\delta}$ follows independent and identical Gaussian distributions, then the Euclidean distance error $\Vert \bm{\delta} \Vert_{2}$ will correspond to a Chi-square distribution\cite{teaser}:
\begin{equation} \label{eq5}
\bm{\delta} \sim N(\mathbf{0}_{3}, \sigma^{2} \mathbf{I}_{3}),
\end{equation}
\begin{equation} \label{eq6}
\frac{1}{\sigma^{2}} \Vert \mathbf{T}^{pri}(\mathbf{x}_{in}^{p}) - \mathbf{x}_{in}^{q} \Vert_{2}^{2} = \frac{1}{\sigma^{2}} \Vert \bm{\delta} \Vert_{2}^{2} \sim \chi^{2}(3),
\end{equation}
where $\mathbf{x}_{in}^{p}$ and $\mathbf{x}_{in}^{q}$ are an inlier correspondence. Consequently, the variance $\sigma^{2}$ characterizes the magnitude of the coordinate difference. Furthermore, $\sigma^{2}$ also describes the Euclidean distance error distribution of the inliers. To estimate $\sigma^{2}$, an intuitive approach is to obtain a pseudo-inlier set from $\mathbf{C}^{vis}$, as inliers typically have much lower coordinate differences compared to outliers:
\begin{equation} \label{eq7}
\mathbf{C}^{in} = \{\mathbf{c}_{i} \in \mathbf{C}^{vis} \mid \Vert \mathbf{T}^{pri}(\mathbf{x}^{p}_{\mathbf{c}_{i}}) - \mathbf{x}^{q}_{\mathbf{c}_{i}} \Vert_{2} \leq t_{inlier} \}.
\end{equation}
Using the pseudo-inlier set $\mathbf{C}^{in}$, we can calculate the distance error and determine the moment estimation of $\sigma^{2}$:
\begin{equation} \label{eq8}
\hat{\sigma}^{2} = \frac{1}{3\vert \mathbf{C}^{in} \vert} \sum_{i}^{\vert \mathbf{C}^{in} \vert}{\Vert \mathbf{T}^{pri}(\mathbf{x}^{p}_{\mathbf{c}_{i}}) - \mathbf{x}^{q}_{\mathbf{c}_{i}} \Vert_{2}^{2}}.
\end{equation}
When $\mathbf{T}^{pri}$ is approximately correct, the estimated $\hat{\sigma}^{2}$ can effectively reflect the confidence.
\par
\noindent \textbf{Distribution-based Search Zone.} To overcome the global ambiguity in geometric feature matching, we design a distribution-based search zone. The goal of the local search zone is to be as small as possible while still encompassing the corresponding point. For each point in the source point cloud $\mathbf{P}$, we use $\mathbf{T}^{pri}$ to map it onto the target point cloud $\mathbf{Q}$. Due to errors in $\mathbf{T}^{pri}$, the transformed point will have a distance error with its corresponding point. Using $\hat{\sigma}^{2}$, we can probabilistically describe the distance error under $\mathbf{T}^{pri}$, e.g., using the quantile $\gamma^{2}$ of the Chi-square distribution with three degrees of freedom and confidence $\alpha$ to find a distance error interval $[0,\epsilon]$ that most of the correct matches are expected to fall in:
\begin{equation} \label{eq9}
P(\frac{1}{\sigma^{2}} \Vert \bm{\delta} \Vert_{2}^{2} \leq \gamma^{2}) = \alpha,\ \epsilon = \hat{\sigma}^{2} \gamma^{2}.
\end{equation}
As a result, we can determine a sphere with the transformed point as its center, and the points within the sphere constitute the local search space of $\mathbf{x}^{p}_{i}$:
\begin{equation} \label{eq10}
\mathbf{Q}^{s}_{i} = \{\mathbf{x}^{q}_{j} \in \mathbf{Q} \mid \Vert \mathbf{T}^{pri}(\mathbf{x}^{p}_{i}) - \mathbf{x}^{q}_{j} \Vert_{2}^{2} \leq \epsilon\}.
\end{equation}
The size of search space $\mathbf{Q}^{s}_{i}$ depends on $\hat{\sigma}^{2}$, making it scale-adaptive and flexible. A larger $\hat{\sigma}^{2}$ results in a larger $\mathbf{Q}^{s}_{i}$, indicating that a rough $\mathbf{T}^{pri}$ requires a large search zone to ensure that the corresponding point of $\mathbf{x}^{p}_{i}$ is within the search space. When $\mathbf{T}^{pri}$ is reliable, the corresponding point will be close to the center of sphere, and the low $\hat{\sigma}^{2}$ can lead to a narrow search zone for removing more irrelevant points. In both cases, the global ambiguity are effectively resolved.
\par
\noindent \textbf{Point-wise Local Match.} After determining the local search space of $\mathbf{x}^{p}_{i}$, we find the nearest neighbor in the feature space as its corresponding point:
\begin{equation} \label{eq11}
\mathbf{C}^{geo'}_{i} = (\mathbf{x}^{p}_{i}, \mathop{argmin}\limits_{\mathbf{x}^{q}_{j} \in \mathbf{Q}^{s}_{i}} \Vert \mathbf{f}^{p}_{i} - \mathbf{f}^{q}_{j} \Vert_{2}^{2}),
\end{equation}
where $\mathbf{f}^{p}_{i}$ and $\mathbf{f}^{q}_{j}$ are the geometric features of $\mathbf{x}^{p}_{i}$ and $\mathbf{x}^{q}_{j}$. Geometric correspondences are extracted by finding the corresponding points in the target point cloud for all source points being searched. If a source point has an empty search space, it will be regarded as a non-overlap point and thus be discarded. With visual guidance, point-wise local match can effectively extract dense and reliable correspondences through geometric features, which are the key to achieving accurate registration. To improve the efficiency of processing large point cloud, we randomly sample a subset of points from the source point cloud $\mathbf{P}$ for search zone determination and local matching, rather than using all points in $\mathbf{P}$. The extracted geometric correspondences and given visual correspondences constitute the final correspondence set $\mathbf{C}$.

\begin{figure}[tp]
    \centering
    \includegraphics[width=\linewidth]{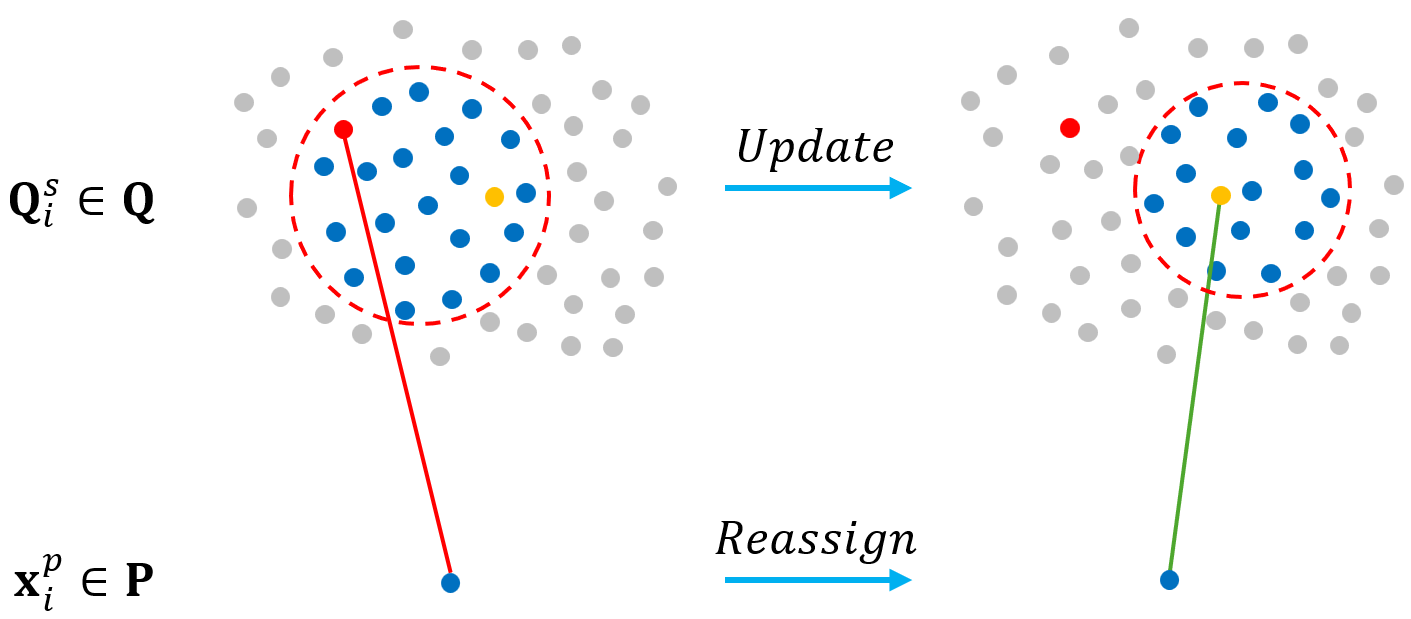}
    \caption{Iterative strategy uses the updated transformation to redetermine clearer local search zones for $\mathbf{x}^{p}_{i}$, which helps reassign the incorrect or ambiguous matches when $\mathbf{T}^{pri}$ is inaccurate.}
    \label{fig:iters}
    \vspace{-1em}
\end{figure}

\subsection{Transformation Estimation}
\noindent \textbf{Geometric Fitting.}
Using the final set $\mathbf{C}$, we solve for the transformation as follows:
\begin{equation} \label{eq12}
\mathbf{T} = \mathop{argmin}\limits_{\mathbf{T}} \sum^{\vert \mathbf{C} \vert}_{i=1}{\omega_{i} \Vert \mathbf{T}(\mathbf{x}_{\mathbf{c}_{i}}^{p}) - \mathbf{x}_{\mathbf{c}_{i}}^{q} \Vert_{2}^{2}},
\end{equation}
where $\omega_{i}$ is the weight, calculated based on the feature similarity between $\mathbf{x}_{\mathbf{c}_{i}}^{p}$ and $\mathbf{x}_{\mathbf{c}_{i}}^{q}$. This can be solved using weighted SVD\cite{dgr}. Since the extracted correspondences are high-quality, the SVD approach can obtain a highly accurate transformation.
\par
\noindent \textbf{Iterative Strategy.} If $\mathbf{T}^{pri}$ is inaccurate, the determined local search zones become unclear, degrading the quality of the extracted correspondences. To relieve the negative effects of an inaccurate $\mathbf{T}^{pri}$ and improve robustness, we design an iterative strategy. After each geometric fitting, we treat the estimated $\mathbf{T}$ as the new $\mathbf{T}^{pri}$, and repeat the visual-guided geometric matching and geometric fitting. For strong matches, the slight movement of local search zone does not affect them. However, for incorrect or ambiguous matches, a more accurate local search zone helps reassign them to the correct points, as shown in \cref{fig:iters}.
\section{Experiment}
\label{sec:experiment}

\begin{table*}[tp]
\centering
\tabcolsep = 6.5pt
\fontsize{8}{10}\selectfont
\begin{tabular}{*{11}{c}}
    \toprule
    \multirow{3}{*}{} & \multicolumn{5}{c}{\textbf{20 frames apart}} & \multicolumn{5}{c}{\textbf{60 frames apart}} \\
    & \multicolumn{2}{c}{Rotation(deg)} & \multicolumn{2}{c}{Translation(cm)} & \multirow{2}{*}{RR} & \multicolumn{2}{c}{Rotation(deg)} & \multicolumn{2}{c}{Translation(cm)} & \multirow{2}{*}{RR} \\
    \cmidrule(lr){2-3} \cmidrule(lr){4-5} \cmidrule(lr){7-8} \cmidrule(lr){9-10}
    & Acc@$2/5/10$ & Err & Acc@$5/10/25$ & Err & & Acc@$2/5/10$ & Err & Acc@$5/10/25$ & Err & \\
    \midrule
    \multicolumn{11}{l}{\textbf{learning-free}} \\
    FPFH+MAC & 90.1 / 98.6 / 99.5 & 0.8 & 84.6 / 95.8 / 98.8 & 2.3 & 99.1 & 59.3 / 85.4 / 90.1 & 1.7 & 52.1 / 77.0 / 87.4 & 4.8 & 87.9 \\
    FPFH+SC$^{2}$-PCR++ & 90.5 / 98.9 / \textbf{99.6} & 0.8 & 84.8 / 96.0 / 99.0 & 2.3 & 99.1 & 59.6 / 88.1 / \textbf{92.4} & 1.7 & 52.3 / 77.2 / 87.8 & 4.8 & 88.1 \\
    Ours(SIFT+FPFH) & \textbf{94.1} / \textbf{99.1} / 99.4 & \textbf{0.6} & \textbf{88.0} / \textbf{96.3} / \textbf{99.3} & \textbf{1.8} & \textbf{99.4} & \textbf{68.6} / \textbf{89.4} / 92.3 & \textbf{1.3} & \textbf{59.3} / \textbf{79.8} / \textbf{90.2} & \textbf{3.9} & \textbf{90.7} \\
    \midrule
    \multicolumn{11}{l}{\textbf{learning-based}} \\
    FCGF+MAC & 91.1 / 98.8 / 99.5 & 0.8 & 85.4 / 95.9 / 99.2 & 2.3 & 99.3 & 63.1 / 88.4 / 92.9 & 1.6 & 55.5 / 79.7 / 90.3 & 4.5 & 91.0 \\
    FCGF+SC$^{2}$-PCR++ & 91.5 / 99.0 / 99.6 & 0.8 & 85.5 / 96.0 / 99.3 & 2.3 & 99.3 & 62.9 / 89.4 / 93.5 & 1.6 & 55.1 / 80.1 / 90.6 & 4.5 & 90.8 \\
    FCGF+VBReg & 91.6 / 98.9 / 99.5 & 0.8 & 85.5 / 96.2 / 99.3 & 2.3 & 99.3 & 62.1 / 88.8 / 92.5 & 1.6 & 55.4 / 80.5 / 90.6 & 4.5 & 90.9 \\
    GeoTransformer & 92.3 / 99.0 / 99.5 & 0.8 & 84.7 / 96.3 / 99.3 & 2.3 & 99.4 & 67.0 / 90.7 / 93.9 & 1.5 & 57.7 / 80.0 / 91.4 & 4.1 & 91.8 \\
    PointMBF & 91.0 / 99.1 / 99.7 & 0.7 & 86.9 / 96.9 / 99.5 & 2.1 & 99.6 & 60.3 / 82.1 / 86.2 & 1.6 & 52.5 / 75.9 / 84.1 & 4.7 & 84.6 \\
    PEAL & 93.0 / 99.1 / 99.7 & 0.8 & 85.5 / 96.6 / 99.6 & 2.3 & 99.6 & 68.5 / 91.7 / 95.0 & 1.4 & 58.1 / 82.2 / 92.3 & 4.0 & 93.1 \\
    Ours(LG+FCGF) & \textbf{94.9} / \textbf{99.6} / \textbf{99.8} & \textbf{0.6} & \textbf{89.0} / \textbf{97.4} / \textbf{99.7} & \textbf{1.8} & \textbf{99.8} & \textbf{72.0} / \textbf{94.3} / \textbf{97.0} & \textbf{1.3} & \textbf{62.6} / \textbf{84.7} / \textbf{94.3} & \textbf{3.7} & \textbf{95.3} \\
    \bottomrule
\end{tabular}
\caption{Registration results on 3DMatch. LG denotes LightGlue. MAC, SC$^{2}$-PCR++ and VBReg are robust estimator methods, GeoTransformer, PointMBF and PEAL are feature learning methods. RGB-D inputs are used for PointMBF, PEAL and our method.}
\label{table1}
\end{table*}

\begin{table*}[tp]
\centering
\tabcolsep = 6.5pt
\fontsize{8}{10}\selectfont
\begin{tabular}{*{11}{c}}
    \toprule
    \multirow{3}{*}{} & \multicolumn{5}{c}{\textbf{20 frames apart}} & \multicolumn{5}{c}{\textbf{60 frames apart}} \\
    & \multicolumn{2}{c}{Rotation(deg)} & \multicolumn{2}{c}{Translation(cm)} & \multirow{2}{*}{RR} & \multicolumn{2}{c}{Rotation(deg)} & \multicolumn{2}{c}{Translation(cm)} & \multirow{2}{*}{RR} \\
    \cmidrule(lr){2-3} \cmidrule(lr){4-5} \cmidrule(lr){7-8} \cmidrule(lr){9-10}
    & Acc@$2/5/10$ & Err & Acc@$5/10/25$ & Err & & Acc@$2/5/10$ & Err & Acc@$5/10/25$ & Err & \\
    \midrule
    \multicolumn{11}{l}{\textbf{learning-free}} \\
    FPFH+MAC & 84.5 / 96.0 / 98.3 & 0.9 & 82.7 / 92.3 / 96.4 & 2.2 & 96.6 & 43.9 / 66.8 / 74.8 & 2.4 & 42.0 / 58.1 / 67.5 & 6.7 & 68.0 \\
    FPFH+SC$^{2}$-PCR++ & 84.8 / \textbf{97.0} / \textbf{99.0} & 0.9 & 81.6 / 92.9 / 96.8 & 2.3 & 97.2 & 43.5 / 68.3 / \textbf{76.2} & 2.4 & 40.0 / 56.9 / 67.4 & 7.2 & 67.9 \\
    Ours(SIFT+FPFH) & \textbf{92.9} / 96.5 / 97.6 & \textbf{0.6} & \textbf{89.2} / \textbf{95.6} / \textbf{97.6} & \textbf{1.5} & \textbf{97.6} & \textbf{56.3} / \textbf{70.3} / 73.6 & \textbf{1.6} & \textbf{50.4} / \textbf{63.6} / \textbf{71.9} & \textbf{4.9} & \textbf{72.3} \\
    \midrule
    \multicolumn{11}{l}{\textbf{learning-based}} \\
    FCGF+MAC & 86.3 / 96.9 / 98.7 & 0.9 & 82.9 / 94.3 / 97.7 & 2.3 & 97.9 & 48.1 / 72.3 / 79.2 & 2.1 & 44.0 / 63.8 / 75.2 & 6.0 & 75.5 \\
    FCGF+SC$^{2}$-PCR++ & 87.0 / 97.1 / 98.8 & 0.9 & 83.6 / 94.8 / 98.0 & 2.3 & 98.1 & 50.2 / 73.8 / 80.4 & 2.0 & 44.6 / 64.9 / 76.0 & 5.8 & 76.6 \\
    FCGF+VBReg & 87.0 / 96.9 / 98.6 & 0.9 & 83.7 / 94.8 / 97.8 & 2.3 & 97.9 & 49.8 / 73.1 / 79.3 & 2.0 & 45.5 / 65.4 / 75.5 & 5.8 & 75.8 \\
    GeoTransformer & 88.3 / 97.0 / 98.3 & 0.8 & 85.6 / 95.1 / 97.7 & 2.1 & 97.9 & 54.1 / 75.6 / 81.1 & 1.8 & 50.4 / 69.1 / 77.5 & 4.9 & 78.4 \\
    PointMBF & 90.8 / 97.8 / 99.0 & 0.7 & 87.6 / 96.8 / 98.9 & 1.9 & 98.8 & 44.9 / 62.6 / 68.4 & 2.4 & 39.2 / 56.0 / 66.3 & 7.5 & 65.9 \\
    PEAL & 89.9 / 97.7 / 98.7 & 0.8 & 87.1 / 95.9 / 98.2 & 2.0 & 98.3 & 58.7 / 79.8 / 84.4 & 1.6 & 53.7 / 73.0 / 81.1 & 4.5 & 81.6 \\
    Ours(LG+FCGF) & \textbf{95.3} / \textbf{99.3} / \textbf{99.7} & \textbf{0.6} & \textbf{91.7} / \textbf{97.9} / \textbf{99.6} & \textbf{1.4} & \textbf{99.7} & \textbf{66.2} / \textbf{85.7} / \textbf{88.9} & \textbf{1.3} & \textbf{59.8} / \textbf{77.7} / \textbf{87.0} & \textbf{3.8} & \textbf{87.4} \\
    \bottomrule
\end{tabular}
\caption{Registration results on ScanNet. The models used for learning-based registration methods are trained on 3DMatch dataset.}
\label{table2}
\vspace{-1em}
\end{table*}

\subsection{Experimental Settings}
\noindent \textbf{Datasets.} We use three datasets that contain both images and point clouds for evaluation. 3DMatch\cite{3dmatch} and KITTI\cite{kitti} datasets are widely used for evaluating geometry-only registration methods, while ScanNet\cite{scannet} dataset has also been adopted by several RGB-D registration studies\cite{byoc, urr, pointmbf} for evaluation. 3DMatch and ScanNet are indoor datasets that use RGB-D sensors to collect color and depth images, whereas KITTI is an outdoor dataset that uses camera and LiDAR to collect images and point clouds.
\par
\noindent \textbf{Metrics.} We use rotation error and translation error to evaluate registration performance\cite{sc2pcr}, report accuracy (Acc) under different error thresholds. Besides, we report registration recall (RR) under a looser error threshold, which is set to ($15^\circ$, $30$ cm) for indoor scenes, and ($5^\circ$, $60$ cm) for outdoor scenes\cite{mac, sc2pcr}. We also report median error (Err), since mean error of whole datas can be greatly influenced by failed pairs and mean error of successfully registered datas cannot reflect the overall accuracy, median error is more referable.
\par
\noindent \textbf{Implementation Details.} We use traditional descriptor FPFH\cite{fpfh} and learning-based descriptor FCGF\cite{fcgf} to extract geometric feature. Besides, FPFH and FCGF are also used to provide initial correspondences for robust estimator methods. SIFT\cite{sift} and LightGlue\cite{lightglue} are used to provide visual matches. For learning-based image matching method LightGlue, we directly use the pretrained model without fine-tuning. 3DMatch, ScanNet and KITTI are all unseen for LightGlue. Additionally, since all benchmark methods except PointMBF\cite{pointmbf} have adopted a common post-refinement approach, we add it to PointMBF for a fair comparison.

\subsection{Results on 3DMatch Dataset}
The 3DMatch and 3DLoMatch datasets used in previous geometry-only methods\cite{mac, sc2pcr, geotransformer} are subsets of the original 3DMatch dataset\cite{3dmatch}, which merge 50 point cloud frames to create registration pairs. However, the process is not suitable for RGB-D methods such as PointMBF and our ViGG, because they are image-dependent, and the merging process is not applicable to images. Therefore, we adopt the setting used in previous RGB-D registration studies\cite{byoc, llt, pointmbf}, selecting color and depth image pairs that are 20 frames apart for evaluation. To make evaluation under lower overlap as in 3DLoMatch, we additionally select pairs that are 60 frames apart. We compare with six recent state-of-the-art methods, including robust estimator methods MAC\cite{mac}, SC$^{2}$-PCR++\cite{sc2pcr++}, and VBReg\cite{vbreg}, and feature learning methods GeoTransformer\cite{geotransformer}, PointMBF\cite{pointmbf}, and PEAL\cite{peal}. Since MAC, SC$^{2}$-PCR++ and our ViGG do not need training, we also make comparisons in a learning-free setting. The results are shown in \cref{table1}, where our method demonstrates the best performance in both learning-free and learning-based settings. Notably, our method shows more improvement in low overlap case. To make further demonstration, we provide an additionally evaluation using pairs that are 120 frames apart in the supplementary material.

\subsection{Results on ScanNet Dataset}
ScanNet\cite{scannet} is a large indoor RGB-D dataset that has been used to evaluate RGB-D registration in previous studies\cite{byoc, llt, pointmbf}. We similarly select pairs that are 20 and 60 frames apart. Following previous studies\cite{byoc, llt, pointmbf}, we use models trained on 3DMatch dataset for learning-based methods and make evaluation on ScanNet test set. As shown in \cref{table2}, our method achieves the best performance in both learning-free and learning-based settings.

\begin{table}[tp]
\centering
\tabcolsep = 1.6pt
\fontsize{8}{10}\selectfont
\begin{tabular}{*{6}{c}}
    \toprule
    \multirow{2}{*}{} & \multicolumn{2}{c}{Rotation(deg)} & \multicolumn{2}{c}{Translation(cm)} & \multirow{2}{*}{RR} \\
    \cmidrule(lr){2-3} \cmidrule(lr){4-5}
    & Acc@$0.25/0.5/1$ & Err & Acc@$7.5/15/30$ & Err & \\
    \midrule
    \multicolumn{6}{l}{\textbf{learning-free}} \\
    FPFH+MAC & 49.4 / 82.5 / 95.7 & 0.25 & 69.5 / 92.3 / 97.5 & 5.9 & 97.8 \\
    FPFH+SC$^{2}$-PCR++ & 55.3 / 85.2 / \textbf{96.0} & 0.23 & 67.2 / 92.6 / \textbf{98.0} & 5.9 & \textbf{98.6} \\
    Ours(SIFT+FPFH) & \textbf{87.7} / \textbf{93.0} / \textbf{96.0} & \textbf{0.08} & \textbf{91.7} / \textbf{94.1} / 96.2 & \textbf{3.1} & 96.8 \\
    \midrule
    \multicolumn{6}{l}{\textbf{learning-based}} \\
    FPFH+VBReg & 52.4 / 84.1 / 95.1 & 0.24 & 64.5 / 91.4 / 97.1 & 6.1 & 97.5 \\
    GeoTransformer & 74.2 / 93.9 / \textbf{97.7} & 0.16 & 77.3 / \textbf{96.4} / \textbf{98.9} & 5.1 & \textbf{99.5} \\
    Ours(LG+FPFH) & \textbf{89.5} / \textbf{94.8} / \textbf{97.7} & \textbf{0.07} & \textbf{92.8} / 96.0 / 98.2 & \textbf{3.0} & 98.9 \\
    \bottomrule
\end{tabular}
\caption{Registration results on KITTI. We use FPFH as the geometric descriptor.}
\label{table3}
\vspace{-1em}
\end{table}

\begin{figure}[tp]
    \centering
    \includegraphics[width=\linewidth]{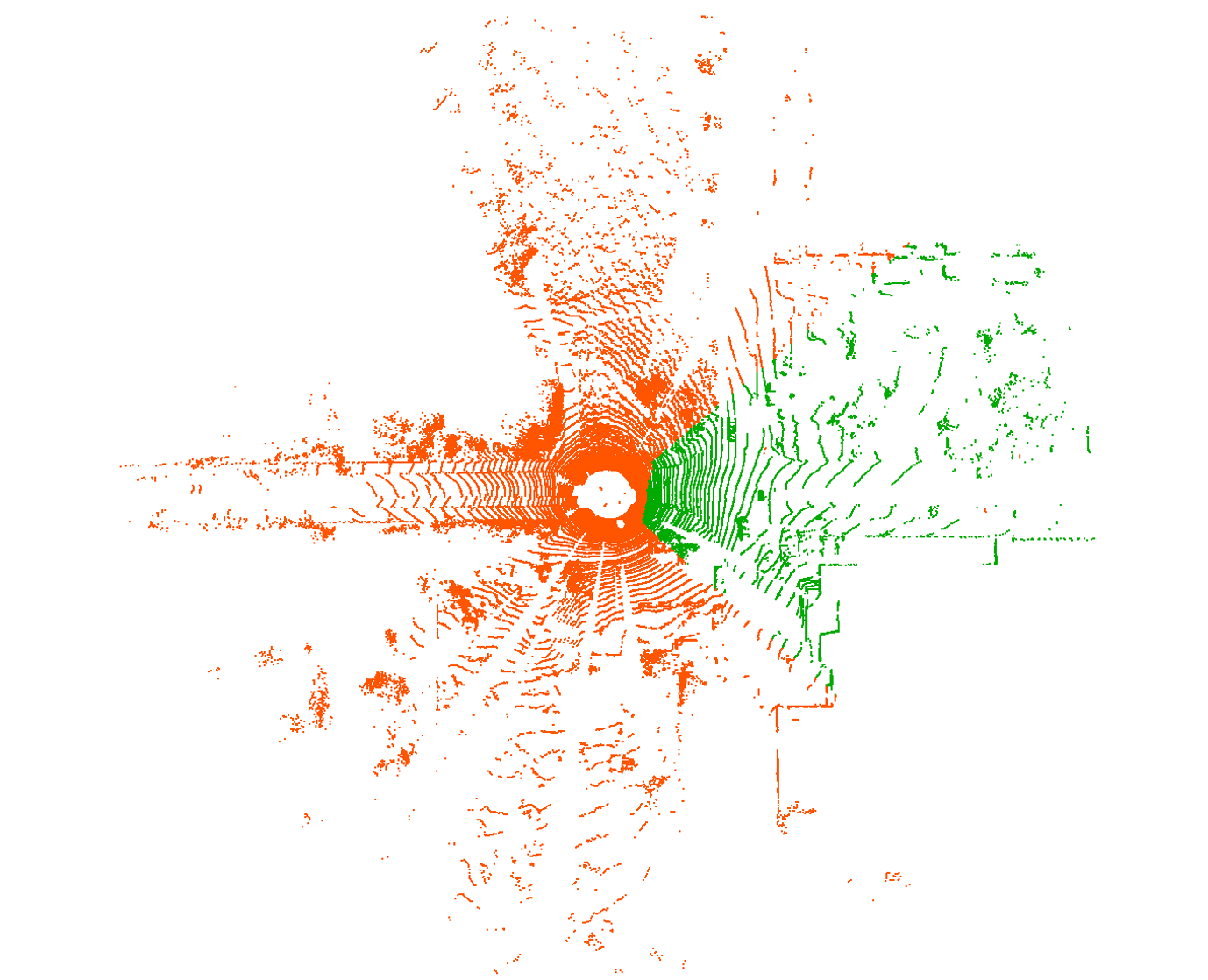}
    \caption{In the KITTI dataset, the camera's field of view is much smaller than that of the LiDAR. The portion of the point cloud within the camera's field of view (colored in green) is less than a quarter of the entire point cloud.}
    \label{fig:kitti}
    \vspace{-1em}
\end{figure}

\subsection{Results on KITTI Dataset}
KITTI\cite{kitti} is an outdoor dataset that includes point clouds collected by LiDAR and color images captured by camera. We follow previous studies\cite{geotransformer, vbreg} to use the same pair splits of sequences 8-10 for test, which contain 555 pairs. We compare with MAC\cite{mac}, SC$^{2}$-PCR++\cite{sc2pcr++}, VBReg\cite{vbreg}, and GeoTransformer\cite{geotransformer}. The results are shown in \cref{table3}, our method achieves significantly lower registration errors. Notably, our method's registration recall is slightly lower than that of other methods. This is because the camera's field of view is much smaller than the LiDAR's, as shown in \cref{fig:kitti}, resulting in a significantly smaller overlap area for visual matching. Therefore, registration using visual matches is considerably more challenging than using geometric matches. In the supplementary material, we provide detailed analysis and evaluation in unified field of view, where our method demonstrates superior improvement.

\subsection{Ablation Studies}
\noindent \textbf{Efficiency of Visual Clique Alignment.} Sparse visual matches can achieve a high inlier ratio, leading to better efficiency compared to geometric matches. To demonstrate this, we adopt the pipeline of MAC without its post-refinement part to evaluate clique alignment performance. As shown in \cref{table4}, SIFT and LightGlue with few correspondences can achieve better accuracy than FPFH and FCGF, and are significantly faster. Besides, we observe that when more than 2500 geometric correspondences are input, there can be tens of GB memory cost difference and considerable time cost difference between low inlier ratio pairs and high inlier ratio pairs. Setting a stricter score threshold to reduce computational costs for high inlier ratio pairs can degrade the performance of low inlier ratio pairs, complicating the balance between accuracy and efficiency. Consequently, the low quantity of visual matches can ensure controllability in the worst case, which is crucial in practical applications.
\begin{table}[tp]
\centering
\tabcolsep = 2pt
\fontsize{8}{10}\selectfont
\begin{tabular}{*{11}{c}}
    \toprule
    \multirow{2}{*}{} & \multicolumn{5}{c}{\textbf{3DMatch60}} & \multicolumn{5}{c}{\textbf{KITTI}} \\
    \cmidrule(lr){2-6} \cmidrule(lr){7-11}
    & Num & RE & TE & RR & TC & Num & RE & TE & RR & TC \\
    \midrule
    SIFT & 150 & 1.6 & 4.7 & 86.3 & 40.8 & 300 & 0.32 & 8.1 & 93.5 & 58.7 \\
    LG & 150 & 1.3 & 4.0 & 94.7 & 72.4 & 300 & 0.29 & 7.6 & 95.9 & 43.5 \\
    \midrule
    \multirow{3}{*}{FPFH} & 300 & 4.1 & 11.8 & 65.2 & 60.5 & 500 & 0.97 & 18.8 & 83.1 & 52.3 \\
    & 500 & 2.9 & 7.9 & 75.4 & 151.3 & 1000 & 0.61 & 12.1 & 93.7 & 220.9 \\
    & 1000 & 2.4 & 6.5 & 81.3 & 473.7 & 2500 & 0.44 & 9.5 & 97.1 & 634.6 \\
    \midrule
    \multirow{3}{*}{FCGF} & 300 & 2.0 & 5.4 & 88.8 & 121.2 & 500 & 0.35 & 17.0 & 95.9 & 182.2 \\
    & 500 & 1.8 & 5.3 & 89.6 & 476.1 & 1000 & 0.32 & 17.2 & 96.0 & 397.4 \\
    & 1000 & 1.7 & 4.9 & 90.4 & 996.6 & 2500 & 0.27 & 16.5 & 96.0 & 1178.1 \\
    \bottomrule
\end{tabular}
\caption{Performance of clique alignment using different matches. Num denotes the number of input correspondences, RE is median rotation error in degree, TE is median translation error in cm, and TC is the average time cost of clique alignment in ms. 3DMatch60 denotes the 3DMatch dataset with 60 frames apart setting.}
\label{table4}
\vspace{-1em}
\end{table}

\begin{table}[tp]
\centering
\tabcolsep = 6.2pt
\fontsize{8}{10}\selectfont
\begin{tabular}{*{8}{c}}
    \toprule
    \multirow{2}{*}{} & \multirow{2}{*}{Setting} & \multicolumn{3}{c}{\textbf{3DMatch60}} & \multicolumn{3}{c}{\textbf{KITTI}} \\
    \cmidrule(lr){3-5} \cmidrule(lr){6-8}
    & & RE & TE & RR & RE & TE & RR \\
    \midrule
    \multirow{5}{*}{SIFT} & no guidance & 1.5 & 4.5 & 86.4 & 0.32 & 7.7 & 93.5 \\
    & w/o GVCA & 1.4 & 4.0 & 88.5 & 0.08 & 3.3 & 94.8 \\
    & w/o VGM & 1.5 & 4.5 & 88.6 & 0.29 & 7.3 & 96.2 \\
    & full & 1.3 & 3.9 & 90.7 & 0.08 & 3.1 & 96.8 \\
    \cmidrule(lr){2-8}
    & no ambiguity & 1.3 & 3.9 & 91.4 & 0.08 & 3.1 & 96.9 \\
    \midrule
    \multirow{5}{*}{LG} & no guidance & 1.3 & 3.8 & 94.7 & 0.29 & 7.0 & 95.9 \\
    & w/o GVCA & 1.3 & 3.7 & 95.2 & 0.07 & 3.0 & 96.8 \\
    & w/o VGM & 1.3 & 3.8 & 94.9 & 0.24 & 6.5 & 98.2 \\
    & full & 1.3 & 3.7 & 95.3 & 0.07 & 3.0 & 98.9 \\
    \cmidrule(lr){2-8}
    & no ambiguity & 1.3 & 3.7 & 95.6 & 0.07 & 3.0 & 98.9 \\
    \bottomrule
\end{tabular}
\caption{Registration performance with different settings. No ambiguity denotes the ideal condition that the ambiguous cliques in visual matches are all removed.}
\label{table5}
\vspace{-1em}
\end{table}
\par
\noindent \textbf{Effectiveness of Mutual Guidance.} To demonstrate the effectiveness of proposed mutual guidance design, we evaluate performance under four configurations: (1) without both geometric-guided visual clique alignment (GVCA) and visual-guided geometric matching (VGM) modules (equivalent to original MAC pipeline), (2) without GVCA module, (3) without VGM module, (4) full ViGG. Additionally, we evaluate the ideal performance without visual ambiguity, achieved by removing all ambiguous cliques using ground truth transformation before hypothesis evaluation. As shown in \cref{table5}, both GVCA and VGM modules effectively improve performance. Notably, ViGG achieves similar results in normal and no ambiguity cases, demonstrating the GVCA module's effectiveness in suppressing ambiguous cliques. 3DMatch uses RGB-D sensors (e.g., Microsoft Kinect, Intel RealSense) to capture RGB-D data with precise 2D-to-3D mappings determined by depth images, which enables image matching method LightGlue to extract sufficiently accurate matches, resulting in less pronounced improvement. However, KITTI uses LiDAR and camera to collect images and point clouds respectively, and the 2D-to-3D mappings cannot be strictly determined. The motion blur, mapping errors, and spatial distribution difference of pixels and 3D points make visual matches noisy. In this case, our method effectively relieves the negative effect brought by noise and achieves significant improvement. To better demonstrate our method's effectiveness, we add Gaussian noise to visual matches to simulate potential noise and mapping errors in practical applications. We set two standard deviations 0.01 and 0.025 to represent low and high noise levels, the results are shown in \cref{table6}. Our method demonstrates significant improvement in these cases. Moreover, compared to non-noisy situation in \cref{table5}, our method exhibits only marginal performance degradation, highlighting its robustness.
\begin{table}[tp]
\centering
\tabcolsep = 7pt
\fontsize{8}{10}\selectfont
\begin{tabular}{*{8}{c}}
    \toprule
    \multirow{2}{*}{} & \multirow{2}{*}{} & \multicolumn{3}{c}{\textbf{Noise 0.01}} & \multicolumn{3}{c}{\textbf{Noise 0.025}} \\
    \cmidrule(lr){3-5} \cmidrule(lr){6-8}
    & & RE & TE & RR & RE & TE & RR \\
    \midrule
    \multirow{2}{*}{SIFT} & MAC & 2.0 & 5.6 & 85.1 & 3.4 & 8.9 & 79.3 \\
    & ViGG(ours) & 1.3 & 4.0 & 90.4 & 1.4 & 4.1 & 88.4 \\
    \midrule
    \multirow{2}{*}{LG} & MAC & 1.5 & 4.3 & 94.4 & 2.4 & 6.5 & 92.5 \\
    & ViGG(ours) & 1.3 & 3.9 & 95.3 & 1.4 & 3.9 & 94.4 \\
    \bottomrule
\end{tabular}
\caption{Registration performance under different noise levels on 3DMatch60. We use two noise versions for evaluation.}
\label{table6}
\vspace{-1em}
\end{table}

\begin{table}[tp]
\centering
\tabcolsep = 11pt
\fontsize{8}{10}\selectfont
\begin{tabular}{*{7}{c}}
    \toprule
    \multirow{2}{*}{Iters} & \multicolumn{2}{c}{\textbf{3DM60}} & \multicolumn{2}{c}{\textbf{N3DM60}} & \multicolumn{2}{c}{\textbf{KITTI}} \\
    \cmidrule(lr){2-3} \cmidrule(lr){4-5} \cmidrule(lr){6-7}
    & RE & TE & RE & TE & RE & TE \\
    \midrule
    w/o & 1.4 & 4.1 & 1.5 & 4.4 & 0.09 & 3.3 \\
    3 & 1.3 & 3.9 & 1.4 & 4.1 & 0.08 & 3.1 \\
    5 & 1.3 & 3.9 & 1.4 & 4.1 & 0.08 & 3.1 \\
    \bottomrule
\end{tabular}
\caption{Registration performance under different iterations. We use SIFT+FPFH to evaluate our method on different datasets. N3DM60 is the 0.025 noise version of 3DMatch60.}
\label{table7}
\vspace{-1em}
\end{table}
\par
\noindent \textbf{Improvement of Iterative Strategy.} We evaluate the registration performance under different iterations to demonstrate the effectiveness of iterative strategy, using 3DMatch60, KITTI and the noise 0.025 version of 3DMatch60 for evaluation. As shown in \cref{table7}, the iterative strategy improves performance. Without any iterations, our method can already achieve remarkable accuracy. However, with iterations, our method can better handle noisy cases and reduce registration errors, thereby achieving better robustness. We find that three iterations are sufficient for convergence, thus the number of iterations is set to three in all experiments.
\par
\noindent \textbf{Time Costs Analysis.} We evaluate the time costs of different registration methods on 3DMatch60 datasets, the results are shown in \cref{table8}. We additionally split the time cost of robust estimator methods and our ViGG into two parts. The experiments are implemented with an Intel i9-10900X CPU and a NVIDIA RTX3090 GPU. The results demonstrate the efficiency of ViGG, indicating that the total time cost of ViGG is lower than that of other methods even if considering the additional time cost of image matching. The high efficiency of ViGG can be attributed to two factors: first, solving clique alignment in the combination form obviously reduces the computational cost; second, the designed VGM module is lightweight and can achieve convergence in few iterations.

\begin{table}[tp]
\centering
\tabcolsep = 5.5pt
\fontsize{8}{12}\selectfont
\begin{tabular}{*{4}{c}}
    \toprule
    & Feature Extraction & Registration & Total \\
    \midrule
    GeoTransformer & - & - & 237.3 \\
    PointMBF & - & - & 221.5 \\
    PEAL & - & - & 1926.4 \\
    FCGF+SC$^{2}$-PCR++ & 65.8 (21.6\%) & 238.3 (78.4\%) & 304.1 \\
    FCGF+VBReg & 65.9 (10.9\%) & 538.4 (89.1\%) & 604.3 \\
    Ours(LG+FCGF) & 112.5 (52.6\%) & 101.2 (47.4\%) & 213.7 \\
    \bottomrule
\end{tabular}
\caption{Average time costs (ms) on 3DMatch60 dataset. Feature Extraction includes the time costs of image matching (only for our ViGG) and geometric feature extraction.}
\label{table8}
\vspace{-1em}
\end{table}
\section{Conclusion}
\label{sec:conclusion}
In this paper, we propose a robust RGB-D point cloud registration method called ViGG. By introducing mutual guidance design, our method effectively leverages the complementary characteristics of visual and geometric features to overcome the limitations. Although effective, our method cannot fundamentally address low-texture scenarios, where the lack of correct visual matches prevents the formation of correct visual cliques. In the future, we would like to explore approaches for enhancing image keypoint matching with geometric information, which could further improve registration performance under low-texture cases.

{
    \small
    \bibliographystyle{ieeenat_fullname}
    \bibliography{main}

@article{teaser,
  title={Teaser: Fast and certifiable point cloud registration},
  author={Yang, Heng and Shi, Jingnan and Carlone, Luca},
  journal={IEEE Transactions on Robotics},
  volume={37},
  number={2},
  pages={314--333},
  year={2020},
  publisher={IEEE}
}

@inproceedings{mac,
  title={3D registration with maximal cliques},
  author={Zhang, Xiyu and Yang, Jiaqi and Zhang, Shikun and Zhang, Yanning},
  booktitle={Proceedings of the IEEE/CVF Conference on Computer Vision and Pattern Recognition},
  pages={17745--17754},
  year={2023}
}

@inproceedings{arcs,
  title={Arcs: Accurate rotation and correspondence search},
  author={Peng, Liangzu and Tsakiris, Manolis C and Vidal, Ren{\'e}},
  booktitle={Proceedings of the IEEE/CVF Conference on Computer Vision and Pattern Recognition},
  pages={11153--11163},
  year={2022}
}

@inproceedings{sc2pcr,
  title={Sc2-pcr: A second order spatial compatibility for efficient and robust point cloud registration},
  author={Chen, Zhi and Sun, Kun and Yang, Fan and Tao, Wenbing},
  booktitle={Proceedings of the IEEE/CVF Conference on Computer Vision and Pattern Recognition},
  pages={13221--13231},
  year={2022}
}

@inproceedings{geotransformer,
  title={Geometric transformer for fast and robust point cloud registration},
  author={Qin, Zheng and Yu, Hao and Wang, Changjian and Guo, Yulan and Peng, Yuxing and Xu, Kai},
  booktitle={Proceedings of the IEEE/CVF conference on computer vision and pattern recognition},
  pages={11143--11152},
  year={2022}
}

@inproceedings{vbreg,
  title={Robust outlier rejection for 3d registration with variational bayes},
  author={Jiang, Haobo and Dang, Zheng and Wei, Zhen and Xie, Jin and Yang, Jian and Salzmann, Mathieu},
  booktitle={Proceedings of the IEEE/CVF conference on computer vision and pattern recognition},
  pages={1148--1157},
  year={2023}
}

@inproceedings{urr,
  title={Unsupervisedr\&r: Unsupervised point cloud registration via differentiable rendering},
  author={El Banani, Mohamed and Gao, Luya and Johnson, Justin},
  booktitle={Proceedings of the IEEE/CVF Conference on Computer Vision and Pattern Recognition},
  pages={7129--7139},
  year={2021}
}

@inproceedings{byoc,
  title={Bootstrap your own correspondences},
  author={El Banani, Mohamed and Johnson, Justin},
  booktitle={Proceedings of the IEEE/CVF International Conference on Computer Vision},
  pages={6433--6442},
  year={2021}
}

@inproceedings{llt,
  title={Improving rgb-d point cloud registration by learning multi-scale local linear transformation},
  author={Wang, Ziming and Huo, Xiaoliang and Chen, Zhenghao and Zhang, Jing and Sheng, Lu and Xu, Dong},
  booktitle={European Conference on Computer Vision},
  pages={175--191},
  year={2022},
  organization={Springer}
}

@inproceedings{pointmbf,
  title={Pointmbf: A multi-scale bidirectional fusion network for unsupervised rgb-d point cloud registration},
  author={Yuan, Mingzhi and Fu, Kexue and Li, Zhihao and Meng, Yucong and Wang, Manning},
  booktitle={Proceedings of the IEEE/CVF International Conference on Computer Vision},
  pages={17694--17705},
  year={2023}
}

@article{sc2pcr++,
  title={SC$^{2}$-PCR++: Rethinking the Generation and Selection for Efficient and Robust Point Cloud Registration},
  author={Chen, Zhi and Sun, Kun and Yang, Fan and Guo, Lin and Tao, Wenbing},
  journal={IEEE Transactions on Pattern Analysis and Machine Intelligence},
  volume={45},
  number={10},
  pages={12358--12376},
  year={2023},
  publisher={IEEE}
}

@inproceedings{peal,
  title={Peal: Prior-embedded explicit attention learning for low-overlap point cloud registration},
  author={Yu, Junle and Ren, Luwei and Zhang, Yu and Zhou, Wenhui and Lin, Lili and Dai, Guojun},
  booktitle={Proceedings of the IEEE/CVF Conference on Computer Vision and Pattern Recognition},
  pages={17702--17711},
  year={2023}
}

@inproceedings{fpfh,
  title={Fast point feature histograms (FPFH) for 3D registration},
  author={Rusu, Radu Bogdan and Blodow, Nico and Beetz, Michael},
  booktitle={2009 IEEE international conference on robotics and automation},
  pages={3212--3217},
  year={2009},
  organization={IEEE}
}

@inproceedings{fcgf,
  title={Fully convolutional geometric features},
  author={Choy, Christopher and Park, Jaesik and Koltun, Vladlen},
  booktitle={Proceedings of the IEEE/CVF international conference on computer vision},
  pages={8958--8966},
  year={2019}
}

@article{sift,
  title={Distinctive image features from scale-invariant keypoints},
  author={Lowe, David G},
  journal={International journal of computer vision},
  volume={60},
  pages={91--110},
  year={2004},
  publisher={Springer}
}

@inproceedings{lightglue,
  title={Lightglue: Local feature matching at light speed},
  author={Lindenberger, Philipp and Sarlin, Paul-Edouard and Pollefeys, Marc},
  booktitle={Proceedings of the IEEE/CVF International Conference on Computer Vision},
  pages={17627--17638},
  year={2023}
}

@inproceedings{3dmatch,
  title={3dmatch: Learning local geometric descriptors from rgb-d reconstructions},
  author={Zeng, Andy and Song, Shuran and Nie{\ss}ner, Matthias and Fisher, Matthew and Xiao, Jianxiong and Funkhouser, Thomas},
  booktitle={Proceedings of the IEEE conference on computer vision and pattern recognition},
  pages={1802--1811},
  year={2017}
}

@inproceedings{scannet,
  title={Scannet: Richly-annotated 3d reconstructions of indoor scenes},
  author={Dai, Angela and Chang, Angel X and Savva, Manolis and Halber, Maciej and Funkhouser, Thomas and Nie{\ss}ner, Matthias},
  booktitle={Proceedings of the IEEE conference on computer vision and pattern recognition},
  pages={5828--5839},
  year={2017}
}

@inproceedings{kitti,
  title={Are we ready for autonomous driving? the kitti vision benchmark suite},
  author={Geiger, Andreas and Lenz, Philip and Urtasun, Raquel},
  booktitle={2012 IEEE conference on computer vision and pattern recognition},
  pages={3354--3361},
  year={2012},
  organization={IEEE}
}

@inproceedings{d3feat,
  title={D3feat: Joint learning of dense detection and description of 3d local features},
  author={Bai, Xuyang and Luo, Zixin and Zhou, Lei and Fu, Hongbo and Quan, Long and Tai, Chiew-Lan},
  booktitle={Proceedings of the IEEE/CVF conference on computer vision and pattern recognition},
  pages={6359--6367},
  year={2020}
}

@inproceedings{icp,
  title={Method for registration of 3-D shapes},
  author={Besl, Paul J and McKay, Neil D},
  booktitle={Sensor fusion IV: control paradigms and data structures},
  volume={1611},
  pages={586--606},
  year={1992},
  organization={Spie}
}

@inproceedings{gicp,
  title={Generalized-icp.},
  author={Segal, Aleksandr and Haehnel, Dirk and Thrun, Sebastian},
  booktitle={Robotics: science and systems},
  volume={2},
  number={4},
  pages={435},
  year={2009},
  organization={Seattle, WA}
}

@article{shot,
  title={SHOT: Unique signatures of histograms for surface and texture description},
  author={Salti, Samuele and Tombari, Federico and Di Stefano, Luigi},
  journal={Computer Vision and Image Understanding},
  volume={125},
  pages={251--264},
  year={2014},
  publisher={Elsevier}
}

@inproceedings{predator,
  title={Predator: Registration of 3d point clouds with low overlap},
  author={Huang, Shengyu and Gojcic, Zan and Usvyatsov, Mikhail and Wieser, Andreas and Schindler, Konrad},
  booktitle={Proceedings of the IEEE/CVF Conference on computer vision and pattern recognition},
  pages={4267--4276},
  year={2021}
}

@inproceedings{colorpcr,
  title={ColorPCR: Color Point Cloud Registration with Multi-Stage Geometric-Color Fusion},
  author={Mu, Juncheng and Bie, Lin and Du, Shaoyi and Gao, Yue},
  booktitle={Proceedings of the IEEE/CVF Conference on Computer Vision and Pattern Recognition},
  pages={21061--21070},
  year={2024}
}

@article{cofinet,
  title={Cofinet: Reliable coarse-to-fine correspondences for robust pointcloud registration},
  author={Yu, Hao and Li, Fu and Saleh, Mahdi and Busam, Benjamin and Ilic, Slobodan},
  journal={Advances in Neural Information Processing Systems},
  volume={34},
  pages={23872--23884},
  year={2021}
}

@inproceedings{ppfnet,
  title={Ppfnet: Global context aware local features for robust 3d point matching},
  author={Deng, Haowen and Birdal, Tolga and Ilic, Slobodan},
  booktitle={Proceedings of the IEEE conference on computer vision and pattern recognition},
  pages={195--205},
  year={2018}
}

@article{ransac,
  title={Random sample consensus: a paradigm for model fitting with applications to image analysis and automated cartography},
  author={Fischler, Martin A and Bolles, Robert C},
  journal={Communications of the ACM},
  volume={24},
  number={6},
  pages={381--395},
  year={1981},
  publisher={ACM New York, NY, USA}
}

@inproceedings{fgr,
  title={Fast global registration},
  author={Zhou, Qian-Yi and Park, Jaesik and Koltun, Vladlen},
  booktitle={Computer Vision--ECCV 2016: 14th European Conference, Amsterdam, The Netherlands, October 11-14, 2016, Proceedings, Part II 14},
  pages={766--782},
  year={2016},
  organization={Springer}
}

@article{rgbdglue,
  title={RGBD-Glue: General Feature Combination for Robust RGB-D Point Cloud Registration},
  author={Chen, Congjia and Jia, Xiaoyu and Zheng, Yanhong and Qu, Yufu},
  journal={arXiv preprint arXiv:2405.07594},
  year={2024}
}

@inproceedings{dgr,
  title={Deep global registration},
  author={Choy, Christopher and Dong, Wei and Koltun, Vladlen},
  booktitle={Proceedings of the IEEE/CVF conference on computer vision and pattern recognition},
  pages={2514--2523},
  year={2020}
}

@inproceedings{pointdsc,
  title={Pointdsc: Robust point cloud registration using deep spatial consistency},
  author={Bai, Xuyang and Luo, Zixin and Zhou, Lei and Chen, Hongkai and Li, Lei and Hu, Zeyu and Fu, Hongbo and Tai, Chiew-Lan},
  booktitle={Proceedings of the IEEE/CVF Conference on Computer Vision and Pattern Recognition},
  pages={15859--15869},
  year={2021}
}

@inproceedings{omniglue,
  title={OmniGlue: Generalizable Feature Matching with Foundation Model Guidance},
  author={Jiang, Hanwen and Karpur, Arjun and Cao, Bingyi and Huang, Qixing and Araujo, Andr{\'e}},
  booktitle={Proceedings of the IEEE/CVF Conference on Computer Vision and Pattern Recognition},
  pages={19865--19875},
  year={2024}
}

@inproceedings{orb,
  title={ORB: An efficient alternative to SIFT or SURF},
  author={Rublee, Ethan and Rabaud, Vincent and Konolige, Kurt and Bradski, Gary},
  booktitle={2011 International conference on computer vision},
  pages={2564--2571},
  year={2011},
  organization={Ieee}
}

@inproceedings{surf,
  title={Surf: Speeded up robust features},
  author={Bay, Herbert and Tuytelaars, Tinne and Van Gool, Luc},
  booktitle={Computer Vision--ECCV 2006: 9th European Conference on Computer Vision, Graz, Austria, May 7-13, 2006. Proceedings, Part I 9},
  pages={404--417},
  year={2006},
  organization={Springer}
}

@article{r2d2,
  title={R2d2: Reliable and repeatable detector and descriptor},
  author={Revaud, Jerome and De Souza, Cesar and Humenberger, Martin and Weinzaepfel, Philippe},
  journal={Advances in neural information processing systems},
  volume={32},
  year={2019}
}

@inproceedings{d2net,
  title={D2-net: A trainable cnn for joint description and detection of local features},
  author={Dusmanu, Mihai and Rocco, Ignacio and Pajdla, Tomas and Pollefeys, Marc and Sivic, Josef and Torii, Akihiko and Sattler, Torsten},
  booktitle={Proceedings of the ieee/cvf conference on computer vision and pattern recognition},
  pages={8092--8101},
  year={2019}
}

@inproceedings{superpoint,
  title={Superpoint: Self-supervised interest point detection and description},
  author={DeTone, Daniel and Malisiewicz, Tomasz and Rabinovich, Andrew},
  booktitle={Proceedings of the IEEE conference on computer vision and pattern recognition workshops},
  pages={224--236},
  year={2018}
}

@inproceedings{superglue,
  title={Superglue: Learning feature matching with graph neural networks},
  author={Sarlin, Paul-Edouard and DeTone, Daniel and Malisiewicz, Tomasz and Rabinovich, Andrew},
  booktitle={Proceedings of the IEEE/CVF conference on computer vision and pattern recognition},
  pages={4938--4947},
  year={2020}
}

@inproceedings{loftr,
  title={LoFTR: Detector-free local feature matching with transformers},
  author={Sun, Jiaming and Shen, Zehong and Wang, Yuang and Bao, Hujun and Zhou, Xiaowei},
  booktitle={Proceedings of the IEEE/CVF conference on computer vision and pattern recognition},
  pages={8922--8931},
  year={2021}
}

@inproceedings{aspanformer,
  title={Aspanformer: Detector-free image matching with adaptive span transformer},
  author={Chen, Hongkai and Luo, Zixin and Zhou, Lei and Tian, Yurun and Zhen, Mingmin and Fang, Tian and Mckinnon, David and Tsin, Yanghai and Quan, Long},
  booktitle={European Conference on Computer Vision},
  pages={20--36},
  year={2022},
  organization={Springer}
}

@inproceedings{kpconv,
  title={Kpconv: Flexible and deformable convolution for point clouds},
  author={Thomas, Hugues and Qi, Charles R and Deschaud, Jean-Emmanuel and Marcotegui, Beatriz and Goulette, Fran{\c{c}}ois and Guibas, Leonidas J},
  booktitle={Proceedings of the IEEE/CVF international conference on computer vision},
  pages={6411--6420},
  year={2019}
}

@inproceedings{spinnet,
  title={Spinnet: Learning a general surface descriptor for 3d point cloud registration},
  author={Ao, Sheng and Hu, Qingyong and Yang, Bo and Markham, Andrew and Guo, Yulan},
  booktitle={Proceedings of the IEEE/CVF conference on computer vision and pattern recognition},
  pages={11753--11762},
  year={2021}
}

@inproceedings{gcransac,
  title={Graph-cut RANSAC},
  author={Barath, Daniel and Matas, Ji{\v{r}}{\'\i}},
  booktitle={Proceedings of the IEEE conference on computer vision and pattern recognition},
  pages={6733--6741},
  year={2018}
}

@inproceedings{3dregnet,
  title={3dregnet: A deep neural network for 3d point registration},
  author={Pais, G Dias and Ramalingam, Srikumar and Govindu, Venu Madhav and Nascimento, Jacinto C and Chellappa, Rama and Miraldo, Pedro},
  booktitle={Proceedings of the IEEE/CVF conference on computer vision and pattern recognition},
  pages={7193--7203},
  year={2020}
}

@inproceedings{3dsc,
  title={Recognizing objects in range data using regional point descriptors},
  author={Frome, Andrea and Huber, Daniel and Kolluri, Ravi and B{\"u}low, Thomas and Malik, Jitendra},
  booktitle={Computer Vision-ECCV 2004: 8th European Conference on Computer Vision, Prague, Czech Republic, May 11-14, 2004. Proceedings, Part III 8},
  pages={224--237},
  year={2004},
  organization={Springer}
}

@inproceedings{sgmnet,
  title={Learning to match features with seeded graph matching network},
  author={Chen, Hongkai and Luo, Zixin and Zhang, Jiahui and Zhou, Lei and Bai, Xuyang and Hu, Zeyu and Tai, Chiew-Lan and Quan, Long},
  booktitle={Proceedings of the IEEE/CVF international conference on computer vision},
  pages={6301--6310},
  year={2021}
}

@inproceedings{regtr,
  title={Regtr: End-to-end point cloud correspondences with transformers},
  author={Yew, Zi Jian and Lee, Gim Hee},
  booktitle={Proceedings of the IEEE/CVF conference on computer vision and pattern recognition},
  pages={6677--6686},
  year={2022}
}

@inproceedings{dynamiccue,
  title={Dynamic Cues-Assisted Transformer for Robust Point Cloud Registration},
  author={Chen, Hong and Yan, Pei and Xiang, Sihe and Tan, Yihua},
  booktitle={Proceedings of the IEEE/CVF Conference on Computer Vision and Pattern Recognition},
  pages={21698--21707},
  year={2024}
}

@inproceedings{tear,
  title={Scalable 3d registration via truncated entry-wise absolute residuals},
  author={Huang, Tianyu and Peng, Liangzu and Vidal, Ren{\'e} and Liu, Yun-Hui},
  booktitle={Proceedings of the IEEE/CVF Conference on Computer Vision and Pattern Recognition},
  pages={27477--27487},
  year={2024}
}

@inproceedings{parenet,
  title={PARE-Net: Position-Aware Rotation-Equivariant Networks for Robust Point Cloud Registration},
  author={Yao, Runzhao and Du, Shaoyi and Cui, Wenting and Tang, Canhui and Yang, Chengwu},
  booktitle={European Conference on Computer Vision},
  pages={287--303},
  year={2024},
  organization={Springer}
}

@article{zhao2023accurate,
  title={Accurate registration of cross-modality geometry via consistent clustering},
  author={Zhao, Mingyang and Huang, Xiaoshui and Jiang, Jingen and Mou, Luntian and Yan, Dong-Ming and Ma, Lei},
  journal={IEEE Transactions on Visualization and Computer Graphics},
  volume={30},
  number={7},
  pages={4055--4067},
  year={2023},
  publisher={IEEE}
}
}

\clearpage
\setcounter{page}{1}
\maketitlesupplementary

\appendix
\setcounter{figure}{0}
\setcounter{table}{0}
\setcounter{equation}{0}
\renewcommand{\thefigure}{S\arabic{figure}}
\renewcommand{\thetable}{S\arabic{table}}

\section{Experimental Setup}
\label{sec:setup}
For 3DMatch and ScanNet datasets, we directly use the depth images provided to generate point clouds, thus the mapping between pixels and 3D points is determined by the depth images. For KITTI dataset, we use the provided calibration parameters to determine the approximate mappings between pixels and 3D points. Specifically, we first transform the point clouds into camera coordinate system using the calibrated transformation matrix between camera and LiDAR, and remove the points that are behind the camera. Then, we project the point clouds onto image plane using the camera intrinsics, and remove the points that are located outside the image. The mapping between pixels and 3D points is then determined by nearest neighbor searching on the image plane.
\par
Following previous works\cite{geotransformer, peal}, we downsample the point cloud with a voxel size of 2.5 cm on indoor datasets 3DMatch and ScanNet, and a voxel size of 30 cm on outdoor dataset KITTI. Besides, the inlier threshold is set to 10 cm for indoor dataets, and 60 cm for outdoor dataset. For the estimated transformation $\mathbf{T}$, the rotation and translation errors are calculated as follows:
\begin{equation} \label{eqs1}
    \mathbf{T} = 
    \left[
    \begin{array}{cc}
        \mathbf{R} & \mathbf{t} \\
        0 & 1 \\
    \end{array}
    \right],
\end{equation}
\begin{equation} \label{eqs2}
\mathrm{
    RE = acos \frac{trace(\mathbf{R}^{T} \bar{\mathbf{R}}) - 1}{2},\ \  TE = \Vert \mathbf{t} - \bar{\mathbf{t}} \Vert_{2},
}
\end{equation}
where $\bar{\mathbf{R}}$ and $\bar{\mathbf{t}}$ are the ground truth rotation matrices and translation vectors.
\par
It should be noted that for the experiments on KITTI dataset, due to the large number of points in the downsampled point cloud, all the robust estimator methods we compared have adopted a subsampling preprocess to control time and memory costs. After feature extraction, they randomly sample $N$ points from the downsampled point cloud before performing feature matching and transformation estimation. In the papers that provide the preprocess code, SC$^{2}$-PCR++\cite{sc2pcr++} randomly samples 8,000 points for evaluation, while VBReg\cite{vbreg} randomly samples 12,000 points for evaluation. However, we observe that this strategy obviously increase the registration errors. This is because randomly sampling a small number of points from a large point cloud increases the distance between points, leading to inaccurate correspondences. As this is not a fair comparison for registration error evaluation, we adjust it to a more appropriate subsampling strategy.
\par
In contrast, we use the entire downsampled point cloud for feature matching, and randomly sample 10,000 correspondences for subsequent estimation to control the computational costs. As shown in \cref{tables1}, our new subsampling strategy does not affect registration recall but achieves lower registration errors. For a fairer and more referable comparison, we apply our new subsampling strategy to all robust estimator methods, and also sample the source points in ViGG's visual-guided geometric matching module to obtain 10,000 correspondences for transformation estimation when evaluating on KITTI dataset.
\par
Furthermore, we observe that using FPFH to provide initial correspondences can achieve better registration performance than FCGF for robust estimator methods on KITTI dataset, which has also been reported in their papers\cite{sc2pcr++, vbreg, mac}. Therefore, we only report the registration performance using FPFH for robust estimator methods and our ViGG on KITTI dataset.

\begin{table}[tp]
\centering
\tabcolsep = 0.9pt
\fontsize{8}{10}\selectfont
\begin{tabular}{*{6}{c}}
    \toprule
    \multirow{2}{*}{} & \multicolumn{2}{c}{Rotation(deg)} & \multicolumn{2}{c}{Translation(cm)} & \multirow{2}{*}{RR} \\
    \cmidrule(lr){2-3} \cmidrule(lr){4-5}
    & Acc@$0.25/0.5/1$ & Err & Acc@$7.5/15/30$ & Err & \\
    \midrule
    SC$^{2}$-PCR++(default) & 44.0 / 79.1 / 95.9 & 0.28 & 55.0 / 90.1 / 97.5 & 7.0 & 98.4 \\
    SC$^{2}$-PCR++ & 55.3 / 85.2 / 96.0 & 0.23 & 67.2 / 92.6 / 98.0 & 5.9 & 98.6 \\
    \midrule
    VBReg(default) & 44.3 / 82.3 / 94.6 & 0.27 & 59.5 / 89.7 / 96.2 & 6.6 & 97.3 \\
    VBReg & 52.4 / 84.1 / 95.1 & 0.24 & 64.5 / 91.4 / 97.1 & 6.1 & 97.5 \\
    \bottomrule
\end{tabular}
\caption{Registration performance with different subsampling strategy on KITTI dataset. FPFH is used to extract features, default denotes using the subsampling strategy provided by their authors. We adjust the subsampling strategy of robust estimator methods to a more appropriate one for a fairer and more referable comparison.}
\label{tables1}
\vspace{-1em}
\end{table}

\section{Detailed Analysis About KITTI Dataset}
\label{sec:kitti}
\begin{figure}[tp]
    \centering
    \includegraphics[width=\linewidth]{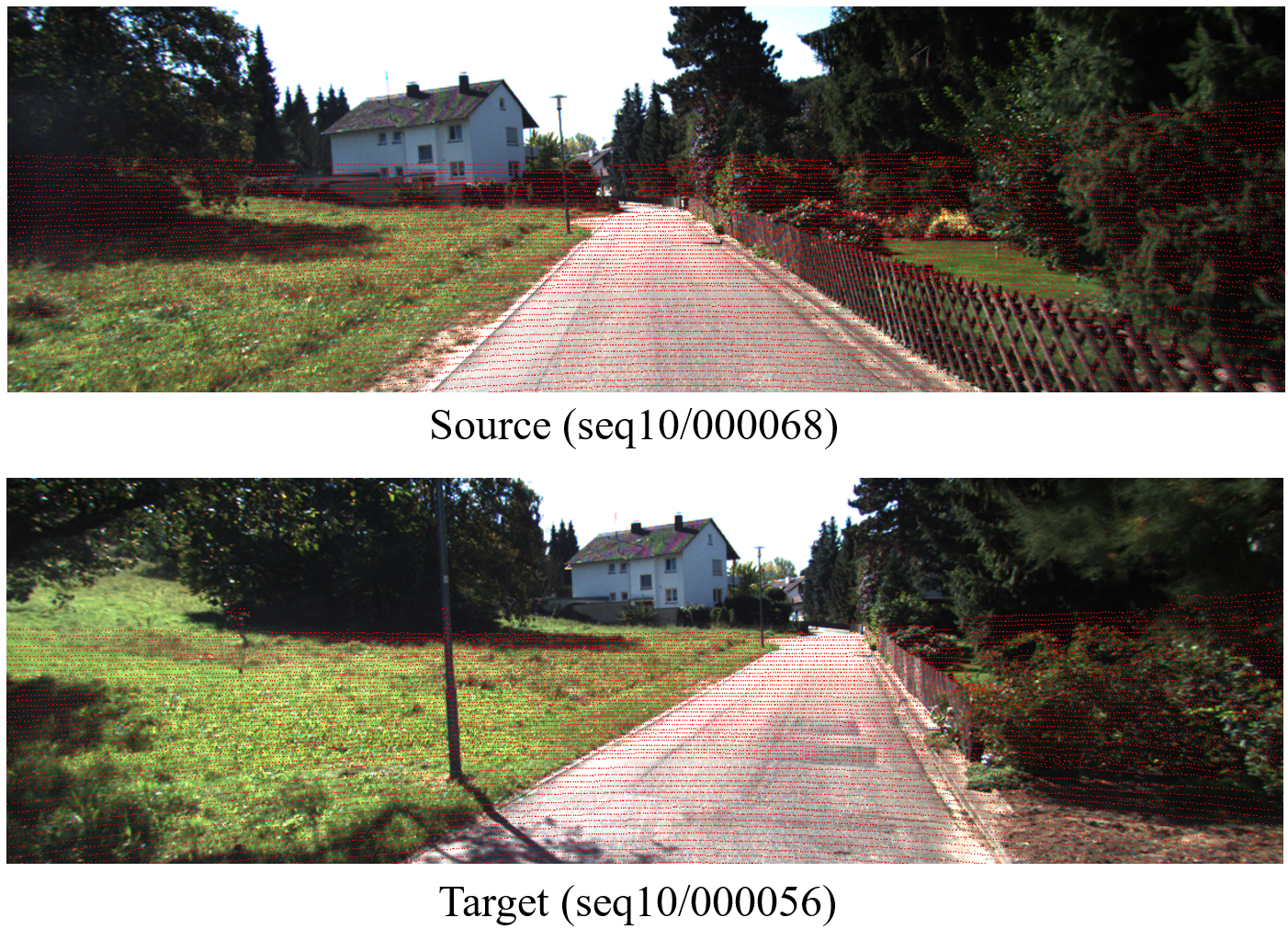}
    \caption{3D points and image pixels are captured by sensors with different ranges, not the entire image is available for registration. We mark the pixels with valid 3D point mapping in red.}
    \label{fig:sup_1}
\end{figure}
\begin{figure}[tp]
    \centering
    \includegraphics[width=\linewidth]{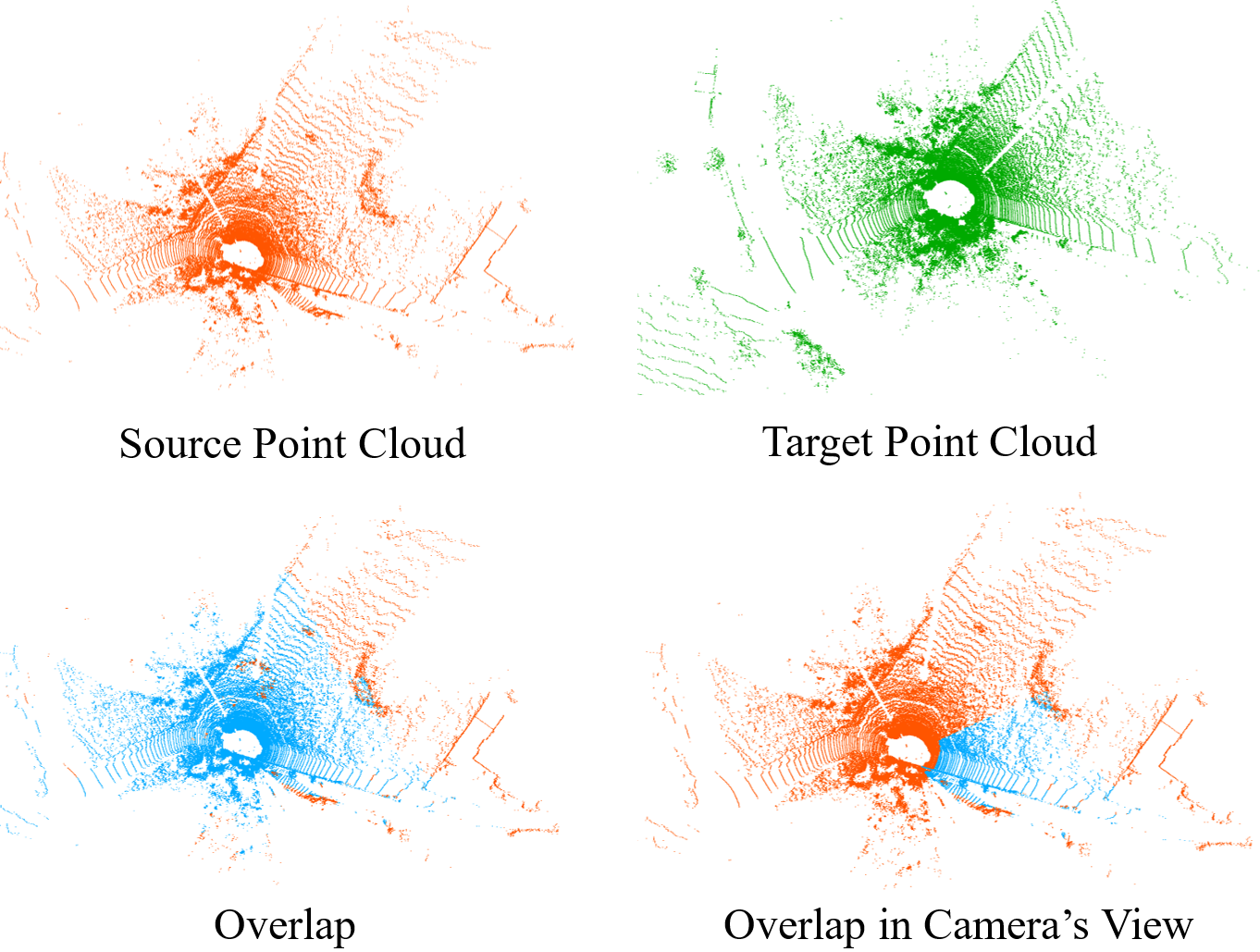}
    \caption{The overlap in the camera's field of view is much smaller than that of the entire point cloud, making registration using visual matches more challenging. We mark the overlapped points of source point cloud in blue.}
    \label{fig:sup_2}
    \vspace{-1em}
\end{figure}
In main paper, we demonstrate that our method achieves significantly lower registration errors compared to other methods. However, our method exhibits slightly lower registration recall. This is primarily because the camera's field of view is much smaller than that of the LiDAR, and as shown in \cref{fig:sup_1}, a quite portion of the image lacks valid mapping to 3D points. Consequently, the valid overlap area between image pairs is small. As shown in \cref{fig:sup_2}, the overlap points within the camera's field of view are much fewer than those in the entire point cloud, leading to less information and fewer potential correspondences, making it unfair for visual matching. KITTI dataset only deploys forward-facing cameras to collect image data, for our proposed ViGG, it is promising that deploying additional cameras in multiple directions to expand the overall field of view can effectively improve visual matching, thereby achieving better registration recall.
\par
\begin{figure}[tp]
    \centering
    \includegraphics[width=\linewidth]{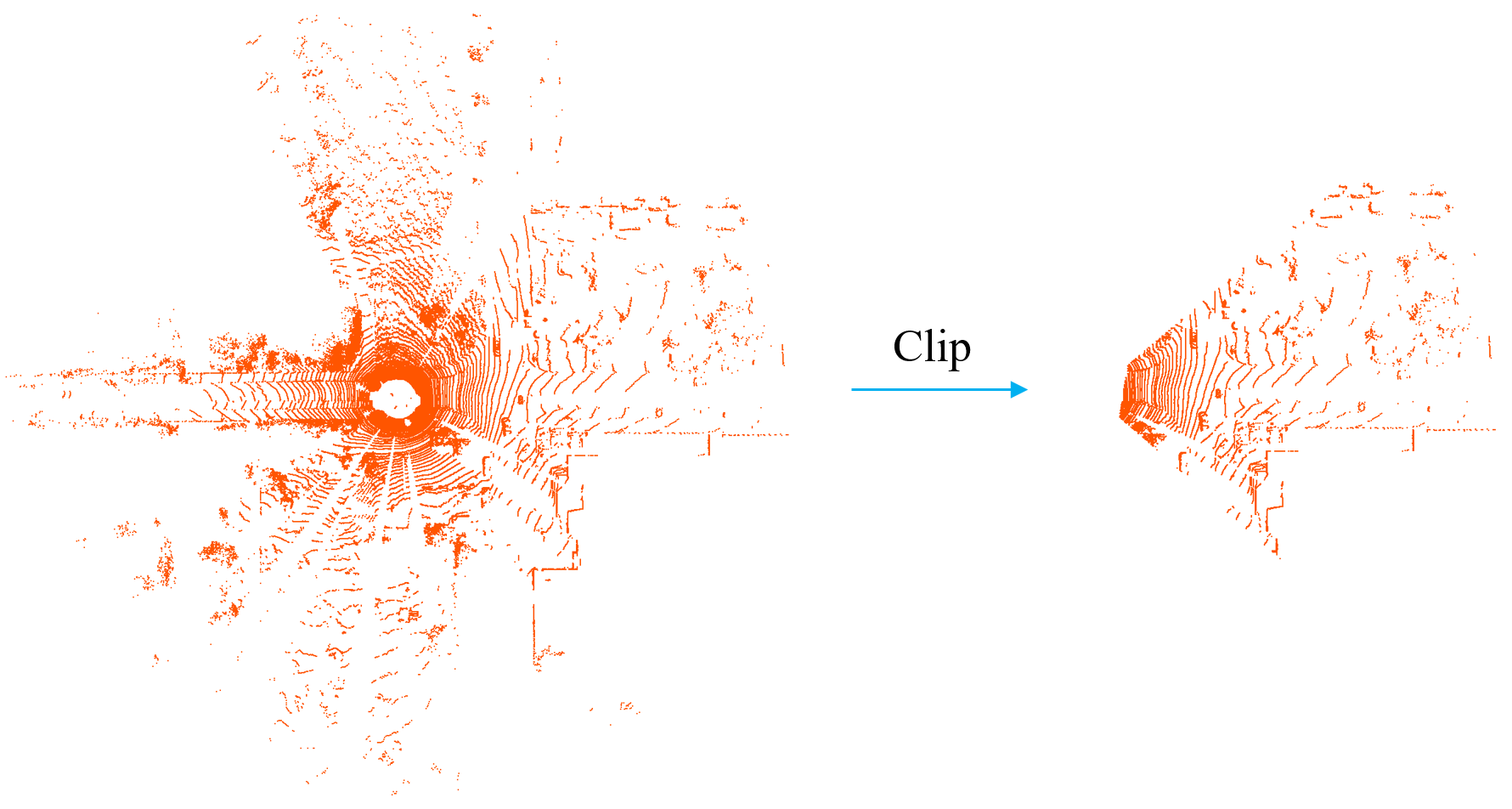}
    \caption{We clip the entire point cloud to retain only the points within the camera's field of view for further experiments.}
    \label{fig:sup_3}
\end{figure}

To further demonstrate our method's effectiveness, we evaluate registration performance within a unified field of view. As shown in \cref{fig:sup_3}, we clip the entire point cloud and retain only the points within the camera's field of view, using the clipped point cloud pairs for registration, which provides a fairer evaluation for visual matches. The results in \cref{tables2} show that in this camera-LiDAR fair case, our method achieves significant improvements in both registration error and recall, highlighting its effectiveness. Additionally, compared to the results in \cref{table3} of main paper, our method achieves better performance when using the entire point cloud for correspondence extraction. This demonstrates that our method can effectively leverage the points without pixel-point mapping, whereas other RGB-D registration methods can only effectively process points that have valid pixel-point mapping.
\begin{table}[tp]
\centering
\tabcolsep = 1.2pt
\fontsize{8}{10}\selectfont
\begin{tabular}{*{6}{c}}
    \toprule
    \multirow{2}{*}{} & \multicolumn{2}{c}{Rotation(deg)} & \multicolumn{2}{c}{Translation(cm)} & \multirow{2}{*}{RR} \\
    \cmidrule(lr){2-3} \cmidrule(lr){4-5}
    & Acc@$0.25/0.5/1$ & Err & Acc@$7.5/15/30$ & Err & \\
    \midrule
    \multicolumn{6}{l}{\textbf{learning-free}} \\
    FPFH+MAC & 14.2 / 44.9 / 72.4 & 0.56 & 24.0 / 55.3 / 75.5 & 13.1 & 79.6 \\
    FPFH+SC$^{2}$-PCR++ & 8.1 / 36.2 / 69.7 & 0.64 & 11.0 / 46.8 / 75.0 & 15.9 & 80.7 \\
    Ours(SIFT+FPFH) & \textbf{74.1} / \textbf{89.2} / \textbf{94.6} & \textbf{0.16} & \textbf{84.9} / \textbf{91.9} / \textbf{94.2} & \textbf{4.0} & \textbf{94.6} \\
    \midrule
    \multicolumn{6}{l}{\textbf{learning-based}} \\
    FPFH+VBReg & 6.8 / 29.4 / 60.5 & 0.76 & 9.5 / 37.1 / 59.3 & 20.5 & 64.9 \\
    GeoTransformer & 44.3 / 74.4 / 89.2 & 0.28 & 49.9 / 80.9 / 85.9 & 7.5 & 86.3 \\
    Ours(LG+FPFH) & \textbf{74.8} / \textbf{91.4} / \textbf{96.2} & \textbf{0.16} & \textbf{86.5} / \textbf{93.7} / \textbf{96.4} & \textbf{3.9} & \textbf{96.9} \\
    \bottomrule
\end{tabular}
\caption{Registration results on KITTI with unified field of view for camera and LiDAR.}
\label{tables2}
\vspace{-1em}
\end{table}

\begin{table}[tp]
\centering
\tabcolsep = 1.6pt
\fontsize{8}{10}\selectfont
\begin{tabular}{*{6}{c}}
    \toprule
    \multirow{2}{*}{} & \multicolumn{2}{c}{Rotation(deg)} & \multicolumn{2}{c}{Translation(cm)} & \multirow{2}{*}{RR} \\
    \cmidrule(lr){2-3} \cmidrule(lr){4-5}
    & Acc@$2/5/10$ & Err & Acc@$5/10/25$ & Err & \\
    \midrule
    \multicolumn{6}{l}{\textbf{learning-free}} \\
    FPFH+MAC & 30.0 / 57.8 / 64.8 & 3.7 & 30.6 / 47.0 / 61.6 & 11.2 & 60.8 \\
    FPFH+SC$^{2}$-PCR++ & 28.6 / 58.2 / 65.4 & 3.6 & 27.8 / 47.4 / 61.0 & 11.2 & 60.6 \\
    Ours(SIFT+FPFH) & \textbf{38.8} / \textbf{63.2} / \textbf{69.4} & \textbf{2.8} & \textbf{35.2} / \textbf{54.8} / \textbf{66.4} & \textbf{8.5} & \textbf{66.4} \\
    \midrule
    \multicolumn{6}{l}{\textbf{learning-based}} \\
    FCGF+MAC & 31.8 / 64.6 / 72.8 & 3.1 & 31.2 / 52.4 / 67.4 & 9.4 & 68.0 \\
    FCGF+SC$^{2}$-PCR++ & 31.6 / 63.6 / 71.6 & 3.1 & 30.8 / 51.6 / 67.0 & 9.1 & 67.4 \\
    FCGF+VBReg & 31.6 / 63.6 / 71.8 & 3.1 & 30.8 / 50.4 / 65.0 & 9.7 & 67.0 \\
    GeoTransformer & 38.2 / 68.8 / 75.0 & 2.7 & 35.2 / 56.4 / 69.6 & 7.9 & 70.6 \\
    PEAL & 40.8 / 70.8 / 79.0 & 2.5 & 36.2 / 59.0 / 73.4 & 7.3 & 75.0 \\
    Ours(LG+FCGF) & \textbf{45.2} / \textbf{77.0} / \textbf{83.8} & \textbf{2.2} & \textbf{41.4} / \textbf{62.0} / \textbf{79.4} & \textbf{6.4} & \textbf{80.4} \\
    \bottomrule
\end{tabular}
\caption{Registration results on 3DMatch dataset with a frame spacing of 120.}
\label{tables3}
\vspace{-1em}
\end{table}

\begin{table*}[tp]
\centering
\tabcolsep = 6.7pt
\fontsize{8}{10}\selectfont
\begin{tabular}{*{17}{c}}
    \toprule
    & \multicolumn{4}{c}{10000} & \multicolumn{4}{c}{5000} & \multicolumn{4}{c}{2500} & \multicolumn{4}{c}{1000} \\
    \cmidrule(lr){2-5} \cmidrule(lr){6-9} \cmidrule(lr){10-13} \cmidrule(lr){14-17}
    & RE & TE & RR & TC & RE & TE & RR & TC & RE & TE & RR & TC & RE & TE & RR & TC \\
    \midrule
    SC$^{2}$-PCR++ & 0.23 & 5.9 & 98.6 & 278.7 & 0.26 & 6.6 & 98.6 & 234.0 & 0.31 & 7.8 & 98.4 & 209.4 & 0.42 & 10.4 & 96.8 & 159.5 \\
    VBReg & 0.24 & 6.1 & 97.5 & 811.7 & 0.27 & 7.2 & 97.5 & 390.6 & 0.34 & 8.1 & 96.4 & 234.1 & 0.52 & 12.0 & 95.3 & 181.5 \\
    Ours(SIFT) & 0.08 & 3.1 & 96.8 & 101.3 & 0.08 & 3.2 & 96.8 & 89.6 & 0.10 & 3.6 & 96.6 & 85.6 & 0.11 & 4.0 & 96.2 & 82.4 \\
    Ours(LG) & 0.07 & 3.0 & 98.9 & 89.7 & 0.08 & 3.1 & 98.7 & 77.6 & 0.10 & 3.3 & 98.4 & 73.6 & 0.11 & 3.9 & 97.7 & 70.1 \\
    \bottomrule
\end{tabular}
\caption{Registration under different sampling settings on KITTI. TC is the total time cost (ms) of the estimation step after feature extraction.}
\label{tables4}
\vspace{-1em}
\end{table*}

\section{Registration under Larger Frame Spacing}
\label{sec:apart}
In main paper, we follow the previous studies to use pairs that are 20 frames apart for registration. Additionally, we use pairs that are 60 frames apart for evaluation under lower overlap. However, we notice that on 3DMatch dataset, 60 frames apart is also easy for our method, making the improvement not obvious. Therefore, we further evaluate registration performance on 3DMatch dataset using pairs that are 120 frames apart, which have extremely low overlap.
\par
As shown in \cref{tables3}, our method has an obvious improvement in learning-free setting, and also outperforms other methods in learning-based setting. Low overlap brings great challenge for feature matching and point cloud registration. However, by effectively leveraging the advantages of both visual and geometric features, our method can better handle low overlap cases and achieve robust registration.

\section{Efficiency Comparison with Robust Estimator Methods}
\label{sec:sampling}
Sampling parts of the correspondences or points for estimation is a commonly used strategy in robust estimator methods, which can reduce the computational cost while bringing only slight accuracy degradation. For registration of large point clouds, sampling process is important. For example, process 10,000 correspondences using VBReg\cite{vbreg} will cost 24 GB GPU memory, making registration computationally expensive.
\par
To further demonstrate our method's efficiency, we evaluate the registration performance and time costs of robust estimator methods and our ViGG under different sampling quantities on KITTI dataset. As mentioned in \cref{sec:setup}, we randomly sample a subset of correspondences to reduce computational costs. SC$^{2}$-PCR++, VBReg and our ViGG are all implemented in Python and run on GPU. As shown in \cref{tables4}, our method achieve both lowest registration errors and time costs in all sampling settings. Notably, ViGG using 10,000 points has already achieved a shorter time cost than SC$^{2}$-PCR++ and VBReg using 1,000 points, indicating that our ViGG is more suitable for handling real-time tasks. Furthermore, our method shows minor accuracy degradation, while SC$^{2}$-PCR++ and VBReg both have a larger accuracy degradation, particularly when sampling 1,000 correspondences. This demonstrates that our method can better balance time cost and accuracy. It should be noted that due to the additional image matching process, our method have extra time costs in the feature extraction step compared to geometry-only methods. However, since the image matching process requires minimal computation time (less than 50 ms on KITTI dataset using LightGlue) and can be further sped up by downsampling the image or choosing a more efficient image matching method, our method still has significant advantages in registration efficiency compared to robust estimator methods.

\section{Comparison of Distribution-based and Fixed Search Zone}
In the visual-guided geometric matching (VGM) module, we propose a distribution-based search zone method to dynamically determine the local search space. This method calculates a radius based on the estimated $\hat{\sigma}^{2}$, allowing the search zone size to adapt to both the noise level and the scale of the point cloud. Benefited from this, the VGM module can achieve optimal matching performance without requiring careful parameter tuning.
\par
To further demonstrate this, we compare the performance of fixed and distribution-based search zones on the noise version of 3DMatch60 dataset and KITTI dataset. For the fixed search zone, we test three different radius values on each dataset. For the distribution-based search zone, we set $\gamma^{2}=10$ across all experiments. As shown in \cref{tables5}, the distribution-based search zone achieves optimal performance on both datasets. In contrast, the fixed search zones suffer performance degradation when improperly configured, requiring dataset-specific parameter tuning. Furthermore, we vary the value of $\gamma^{2}$ to our method's sensitivity. As shown in \cref{tables6}, obvious performance degradation only occurs when $\gamma^{2}$ is low (when set to 2), which may makes the search zone too small and causes the loss of corresponding points. When $\gamma^{2}$ is large, varying its value has almost no effect on registration performance in both two datasets. This indicates that our method is insensitive to the hyper-parameter $\gamma^{2}$.
\par
These results demonstrate the strong generalizability of the distribution-based search zone, which is a significant advantage in practical applications. Notably, when encountering varying noise levels across data, determining suitable search zones with fixed radius becomes impractical, whereas the distribution-based search zone adapts effectively in this case.

\begin{table}[tp]
\centering
\tabcolsep = 6.9pt
\fontsize{8}{10}\selectfont
\begin{tabular}{*{8}{c}}
    \toprule
    \multirow{2}{*}{} & \multirow{2}{*}{Radius} & \multicolumn{3}{c}{\textbf{SIFT}} & \multicolumn{3}{c}{\textbf{LG}} \\
    \cmidrule(lr){3-5} \cmidrule(lr){6-8}
    & & RE & TE & RR & RE & TE & RR \\
    \midrule
    \multicolumn{8}{l}{\textbf{N3DM60}} \\
    \multirow{3}{*}{fixed} & 0.05 & 1.5 & 4.5 & 87.9 & 1.4 & 4.2 & 94.4 \\
    & 0.1 & 1.5 & 4.3 & 88.4 & 1.4 & 4.0 & 94.4 \\
    & 0.2 & 1.6 & 4.5 & 88.3 & 1.5 & 4.1 & 94.2 \\
    \midrule
    dynamic & - & 1.4 & 4.1 & 88.4 & 1.4 & 3.9 & 94.4 \\
    \midrule
    \multicolumn{8}{l}{\textbf{KITTI}} \\
    \multirow{3}{*}{fixed} & 0.3 & 0.08 & 3.6 & 96.6 & 0.08 & 3.6 & 98.6 \\
    & 0.6 & 0.08 & 3.1 & 96.8 & 0.08 & 3.0 & 98.7 \\
    & 1.2 & 0.12 & 3.3 & 96.9 & 0.12 & 3.3 & 98.7 \\
    \midrule
    dynamic & - & 0.08 & 3.1 & 96.8 & 0.07 & 3.0 & 98.9 \\
    \bottomrule
\end{tabular}
\caption{Registration performance using fixed and distribution-based (dynamic) search zone. Radius is measured in m.}
\label{tables5}
\end{table}

\begin{table}[tp]
\centering
\tabcolsep = 10pt
\fontsize{8}{10}\selectfont
\begin{tabular}{*{7}{c}}
    \toprule
    \multirow{2}{*}{$\gamma^{2}$} & \multicolumn{3}{c}{\textbf{SIFT}} & \multicolumn{3}{c}{\textbf{LG}} \\
    \cmidrule(lr){2-4} \cmidrule(lr){5-7}
    & RE & TE & RR & RE & TE & RR \\
    \midrule
    \multicolumn{7}{l}{\textbf{N3DM60}} \\
    2 & 1.5 & 4.2 & 87.6 & 1.4 & 4.1 & 94.4 \\
    5 & 1.4 & 4.1 & 88.6 & 1.4 & 3.9 & 94.4 \\
    10 & 1.4 & 4.1 & 88.4 & 1.4 & 3.9 & 94.4 \\
    20 & 1.4 & 4.2 & 88.4 & 1.4 & 4.0 & 94.5 \\
    \midrule
    \multicolumn{7}{l}{\textbf{KITTI}} \\
    2 & 0.10 & 4.1 & 96.8 & 0.09 & 3.6 & 98.6 \\
    5 & 0.08 & 3.3 & 96.8 & 0.07 & 3.1 & 98.9 \\
    10 & 0.08 & 3.1 & 96.8 & 0.07 & 3.0 & 98.9 \\
    20 & 0.08 & 3.1 & 96.9 & 0.07 & 3.0 & 98.9 \\
    \bottomrule
\end{tabular}
\caption{Registration performance using different values of $\gamma^{2}$.}
\label{tables6}
\vspace{-1em}
\end{table}

\section{Comparison with ICP}
\label{sec:icp}
ICP\cite{icp} is a commonly used registration method when a prior transformation is provided. After clique-based alignment, performing fine registration with ICP using the prior transformation and point cloud pairs is also a feasible approach for RGB-D registration tasks. However, ICP is sensitive to the accuracy of prior transformation and can easily fall into local optima, making it less robust. In contrast, our proposed visual-guided geometric matching (VGM) method can better handle visually noisy cases and achieve significantly more robust and accurate registration.
\par
To make further demonstration, we first use the geometric-guided visual clique alignment (GVCA) module to obtain prior transformation, then evaluate the post estimation performance of post-refinement, ICP and our VGM with SVD. As shown in \cref{tables7}, our method achieve the best performance. On the KITTI dataset where the prior transformation is relatively accurate, our method shows a clear improvement over ICP. When dealing with less accurate prior transformations on N3DM60 dataset, ICP exhibits even worse accuracy than geometry-only methods, and is significantly worse than our method. In this visually noisy case, the inaccurate prior transformations bring great challenge to ICP. However, our VGM can obtain reliable correspondences and achieve accurate registration even when visual matches are noisy, which is crucial for handling various RGB-D registration tasks.

\begin{table}[tp]
\centering
\tabcolsep = 11pt
\fontsize{8}{10}\selectfont
\begin{tabular}{*{6}{c}}
    \toprule
    \multirow{2}{*}{} & \multirow{2}{*}{Post Method} & \multicolumn{2}{c}{N3DM60} & \multicolumn{2}{c}{KITTI} \\
    \cmidrule(lr){3-4} \cmidrule(lr){5-6}
    & & RE & TE & RE & TE \\
    \midrule
    \multirow{3}{*}{SIFT} & Post Refi & 2.8 & 7.6 & 0.29 & 7.3 \\
    & ICP & 1.8 & 6.0 & 0.15 & 3.9 \\
    & VGM+SVD & 1.4 & 4.1 & 0.08 & 3.1 \\
    \midrule
    \multirow{3}{*}{LG} & Post Refi & 2.1 & 5.7 & 0.24 & 6.5 \\
    & ICP & 1.7 & 5.8 & 0.14 & 3.9 \\
    & VGM+SVD & 1.4 & 3.9 & 0.07 & 3.0 \\
    \bottomrule
\end{tabular}
\caption{Post estimation performance using the prior transformation estimated by GVCA. Post Refi denotes applying post-refinement with visual matches. N3DM60 denotes the noise 0.025 version of 3DMatch60.}
\label{tables7}
\end{table}

\section{Issues About Comparing with ColorPCR}
\label{sec:colorpcr}

\begin{table}[tp]
\centering
\tabcolsep = 3.2pt
\fontsize{8}{10}\selectfont
\begin{tabular}{*{6}{c}}
    \toprule
    \multirow{2}{*}{} & \multicolumn{2}{c}{Rotation(deg)} & \multicolumn{2}{c}{Translation(cm)} & \multirow{2}{*}{RR} \\
    \cmidrule(lr){2-3} \cmidrule(lr){4-5}
    & Acc@$2/5/10$ & Err & Acc@$5/10/25$ & Err & \\
    \midrule
    \multicolumn{6}{l}{\textbf{20 frames apart}} \\
    GeoTransformer & 92.3 / 99.0 / 99.5 & 0.8 & 84.7 / 96.3 / 99.3 & 2.3 & 99.4 \\
    ColorPCR & 84.0 / 95.9 / 97.6 & 0.9 & 77.7 / 92.0 / 96.7 & 2.7 & 97.1 \\
    \midrule
    \multicolumn{6}{l}{\textbf{60 frames apart}} \\
    GeoTransformer & 67.0 / 90.7 / 93.9 & 1.5 & 57.7 / 80.0 / 91.4 & 4.1 & 91.8 \\
    ColorPCR & 50.1 / 75.9 / 83.0 & 2.0 & 45.6 / 66.8 / 78.8 & 5.7 & 79.5 \\
    \bottomrule
\end{tabular}
\caption{Registration performance of ColorPCR and GeoTransformer on 3DMatch dataset.}
\label{tables8}
\vspace{-1em}
\end{table}

\begin{figure}[tp]
    \centering
    \includegraphics[width=\linewidth]{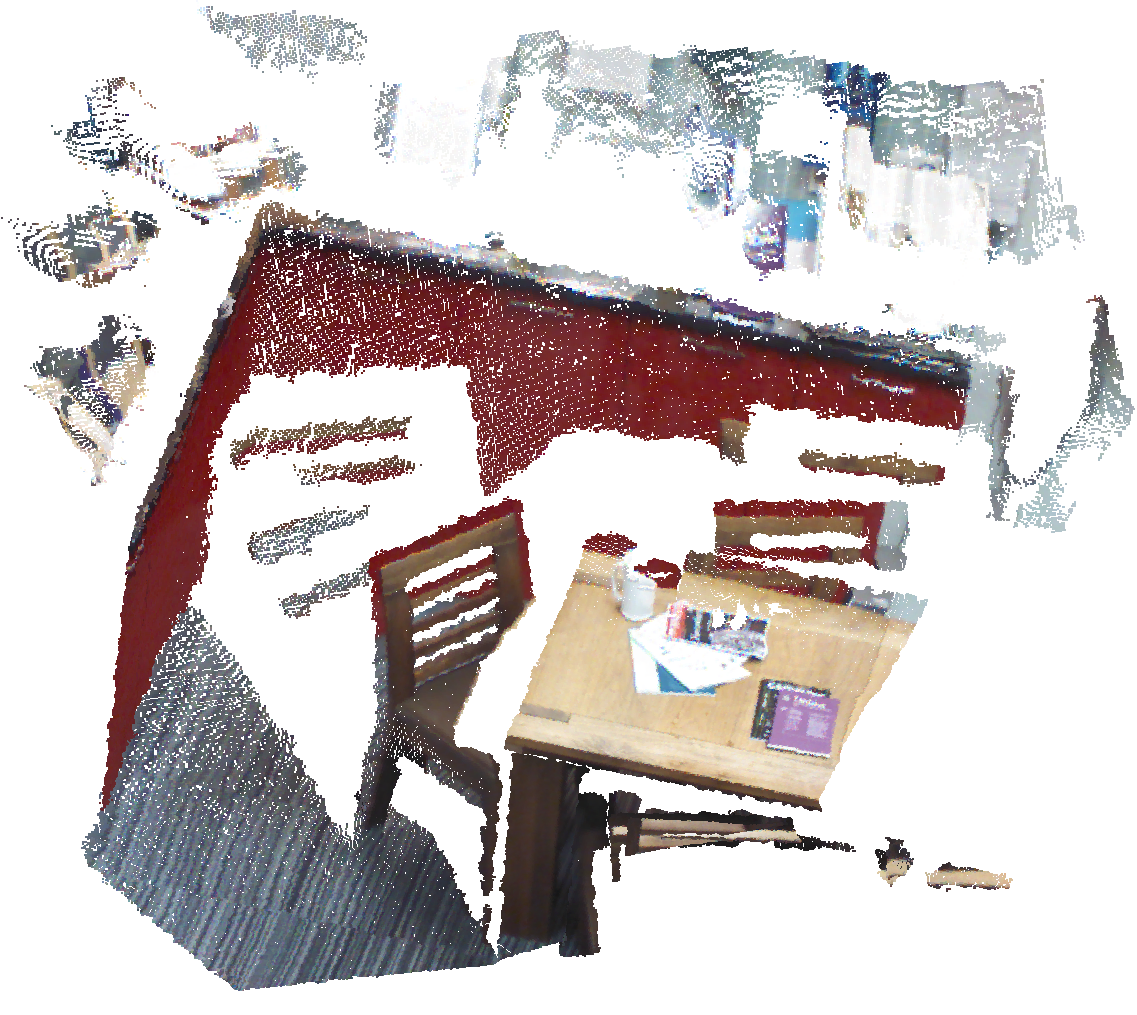}
    \caption{Color point cloud generated from 3DMatch dataset. We generate the color point clouds directly using the color and depth images for ColorPCR.}
    \label{fig:sup_4}
    \vspace{-1em}
\end{figure}

ColorPCR\cite{colorpcr} is the newest RGB-D registration method, which uses color information to enhance GeoTransformer networks and extract more distinctive superpoint features for matching. We reproduce the remarkable performance of ColorPCR on their provided color point cloud datasets Color3DMatch and Color3DLoMatch using the code and pretrained model. However, Color3DMatch and Color3DLoMatch datasets are also 50 franes-merged, which cannot be used for RGB-D registration methods as said in main paper. Therefore, we test ColorPCR on the RGB-D 3DMatch benchmark used in main paper. Unexpectedly, we observe serious performance degradation. As shown in \cref{tables8}, ColorPCR achieves worse registration performance than GeoTransformer\cite{geotransformer}.
\par
After checking the code and discussing with the authors, we suppose the issue lies in the input color point cloud data. ColorPCR first reconstructs the entire scene using the color and depth images from original 3DMatch dataset, then applies the scene’s color to the merged point cloud for generating color point cloud data. However, we follow PointMBF\cite{pointmbf} to process 3DMatch dataset for RGB-D registration evaluation, the registration is performed on frames that are 20 or 60 frames apart, with only a pair of color and depth images available for each registration process. As shown in \cref{fig:sup_4}, we use a frame of color and depth images to directly generate the color point cloud, which may introduce more color errors and negatively affect the performance of ColorPCR. As a result, the RGB-D 3DMatch benchmark used in our main paper may not be suitable for evaluating ColorPCR, and therefore, we do not include ColorPCR in the main paper's comparison.

\end{document}